\documentclass[10pt,twocolumn,letterpaper]{article}

\usepackage[pagenumbers]{cvpr}              %

\definecolor{cvprblue}{rgb}{0.21,0.49,0.74}
\usepackage[pagebackref,breaklinks,colorlinks,citecolor=cvprblue]{hyperref}

\usepackage[utf8]{inputenc} %
\usepackage[T1]{fontenc}    %
\usepackage{enumitem}
\definecolor{bluelink}{rgb}{0.12,0.49,0.85}
\usepackage{url}            %
\usepackage{booktabs}       %
\usepackage{amsfonts}       %
\usepackage{nicefrac}       %
\usepackage{microtype}      %

\usepackage{arydshln}
\usepackage{colortbl}
\usepackage[font=small]{caption}
\usepackage{multirow} %
\usepackage{makecell}
\usepackage{etoolbox}
\usepackage{pgf}
\usepackage{adjustbox}
\usepackage{graphicx}
\usepackage{tabularx}
\usepackage{xspace}
\usepackage{subcaption}
\usepackage{wrapfig}
\usepackage{comment}
\usepackage{amssymb}
\usepackage{float}

\usepackage{listings}
\usepackage{seqsplit}

\usepackage[T1]{fontenc}
\usepackage{fix-cm}

\let\oldparagraph\paragraph
\renewcommand{\paragraph}[1]{%
  \vspace{-10pt}\oldparagraph{#1}%
}

\title{Community Forensics: Using Thousands of Generators to Train \\Fake Image Detectors}

\newcommand{\supparxiv}[2]{#2} %

\author{Jeongsoo Park \qquad\qquad Andrew Owens\\
University of Michigan\\
}

\begin{document}
\maketitle

\newcommand{\totalImages}{2.7M} %
\newcommand{\hfModelNum}{4763}
\newcommand{\hfImagesAvg}{403}
\newcommand{\hfImagesTotal}{1.9M}%
\newcommand{\manualModelNum}{19}
\newcommand{\manualImagesAvg}{40,738}
\newcommand{\manualImagesTotal}{774K}%
\newcommand{\testImagesTotal}{26K} %
\newcommand{\RAISEsampledModels}{8}
\newcommand{\totalModelsInclTest}{4803} %
\newcommand{\totalModelsOnlyTrain}{4782}
\newcommand{\evalModelsOOD}{10} %
\newcommand{\evalImagesOOD}{11K} %
\newcommand{\evalImagesOODEach}{1K}
\newcommand{\evalModelsUnseen}{21} %
\newcommand{\evalImagesUnseen}{26K} %
\newcommand{\commercialImages}{15K} %
\newcommand{\commercialModels}{11}
\newcommand{\datasetName}{{Community Forensics}}
\newcommand{\figvspace}{\vspace{\supparxiv{-4mm}{-4mm}}}
\newcommand{\secVspace}{\vspace{-1mm}}
\newcommand{\subsecVspace}{\vspace{-1mm}}

\newcommand{\modelNum}{\totalModelsInclTest{}}
\newcommand{\modelSizeMult}{34} %

\newcommand{\urlsize}{\fontsize{7.85pt}{9.3pt}\selectfont}

\begin{abstract}

One of the key challenges of detecting AI-generated images is spotting images that have been created by previously unseen generative models. 
We argue that the limited diversity of the training data is a major obstacle to addressing this problem, and we propose a new dataset that is significantly larger and more diverse than prior works. 
As part of creating this dataset, 
we systematically download thousands of text-to-image latent diffusion models and sample images from them. 
We also collect images from dozens of popular open source and commercial models. The resulting dataset contains \totalImages{} images that have been sampled from \modelNum{} different models. These images collectively capture a wide range of scene content, generator architectures, and image processing settings. 
Using this dataset, we study the generalization abilities of fake image detectors. 
Our experiments suggest that detection performance improves as the number of models in the training set increases, even when these models have similar architectures.
We also find that increasing the diversity of the models improves detection performance, 
and that our trained detectors generalize better than those trained on other datasets.
The dataset can be found in \texttt{\textup{\urlsize\url{https://jespark.net/projects/2024/community_forensics}}}

\end{abstract}

\vspace{-1mm}
\section{Introduction}
\label{sec:intro}
\vspace{-1mm}
\begin{figure}[t]
  \begin{center}
    \includegraphics[width=0.45\textwidth]{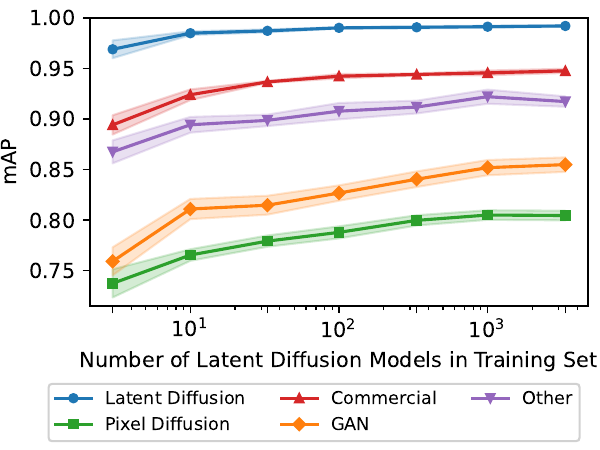}
  \end{center}
  \vspace{-5mm}
    \caption{\small {\bf Performance \vs{} model diversity.} We use images sampled from different numbers of open source latent diffusion models in our {\em Community Forensics} dataset to train fake image detectors 
    (shown in \cref{fig:teaser_systematic}). We fix the total number of images and only vary the number of models.
    The detector's performance increases 
    across all generator types 
    as we train from more models, 
    even though these %
    added models are entirely latent diffusion.
    This improvement is largest for test images from out-of-distribution 
    generators, 
    such as pixel-based diffusion models or GANs.  
    }\vspace{-4mm}
  \label{fig:nummodels} %
\end{figure}

Our ability to automatically generate realistic images is quickly outpacing our ability to detect them, potentially leading to a state of affairs in which neither humans nor machines can reliably tell real from fake. While the field of image forensics has been developing methods to address this problem, existing fake image detectors still struggle with generalization.
These methods often excel at detecting images from generators that were present in their training sets, but fail when given images sampled from unseen models~\cite{rossler2019faceforensics++,Wang_2020_CVPR,Ojha_2023_CVPR,rossler2018faceforensics}.

A core challenge is dealing with the large amounts of variation between models. Each generator has a potentially unique combination of the architecture, loss function, and training distribution. Even seemingly minor differences in low-level image processing details, such as the ways that training images are resized or compressed, can strongly influence detection accuracy~\cite{Wang_2020_CVPR}. As a result of these model-specific idiosyncrasies, a generator's images may evade detection, even when images from architecturally similar models exist in the training set. This issue has been exacerbated by the thousands of open source models that are now available online, many of which extend pretrained base models in complex ways.

\begin{figure*}[t] 
    \centering
        \centering
        \begin{subfigure}[b]{0.6\linewidth}
            \centering
            \includegraphics[width=\linewidth]{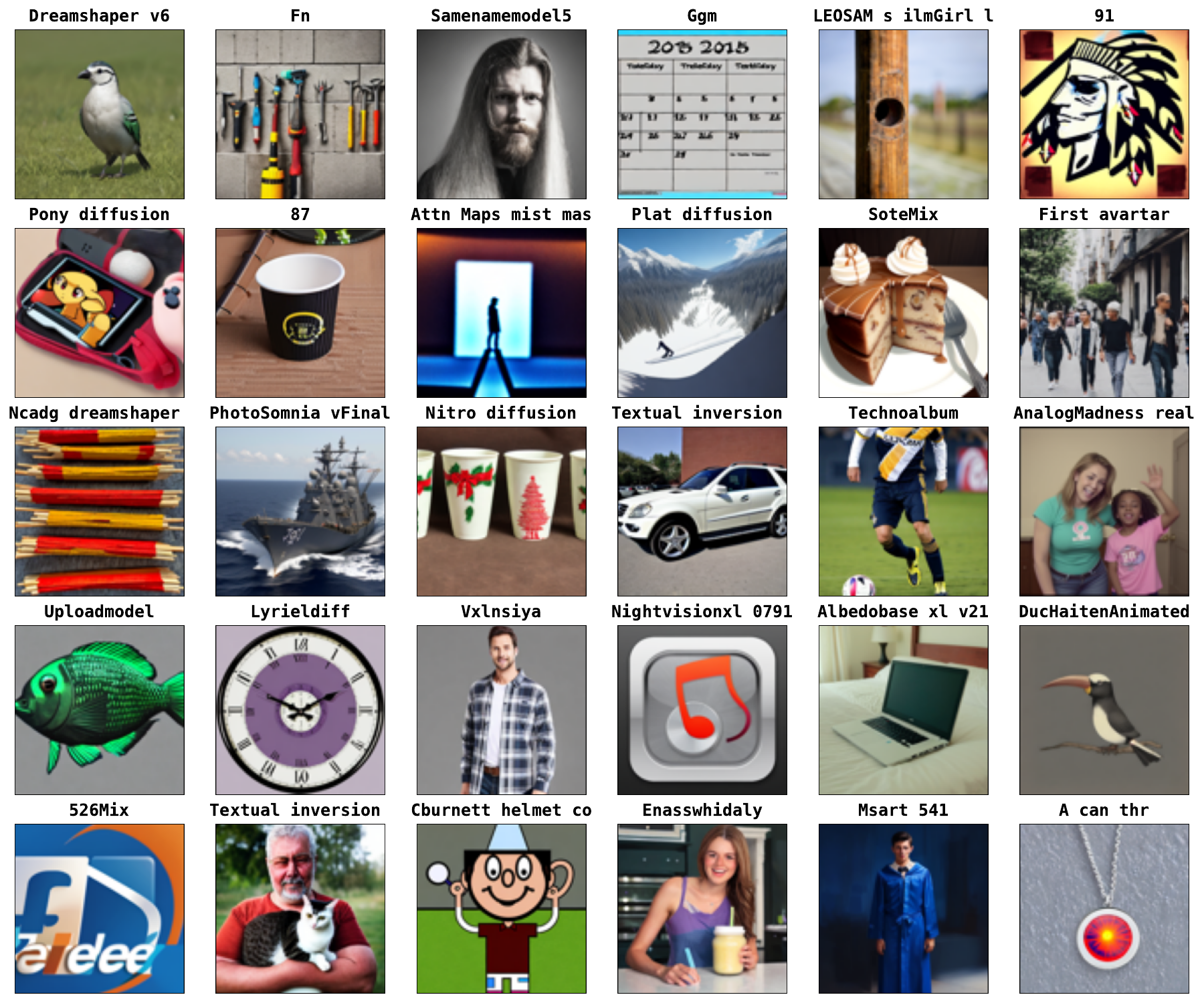}
            \caption{Systematically collected diffusion models (\hfModelNum~models)}
            \label{fig:teaser_systematic}
        \end{subfigure}%
        \hfill
        \begin{subfigure}[b]{0.385\linewidth}
            \centering
            \includegraphics[width=\linewidth]{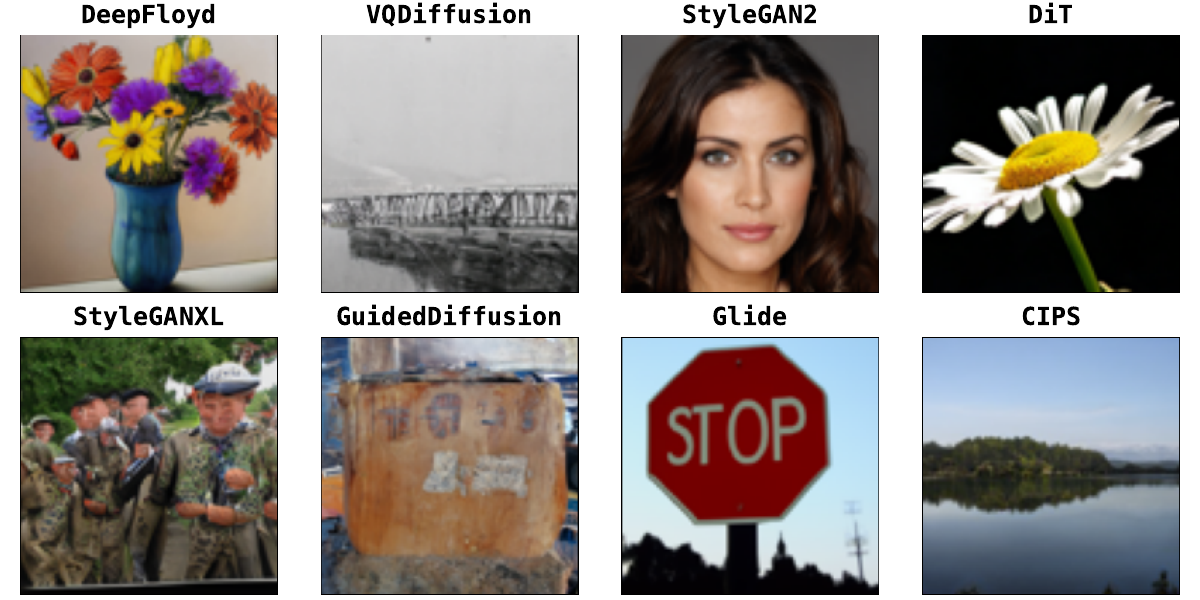}
            \caption{Manually chosen open source models (\manualModelNum~models)}
            \label{fig:teaser_manual}
            \vspace{0.6cm} %
            \includegraphics[width=\linewidth]{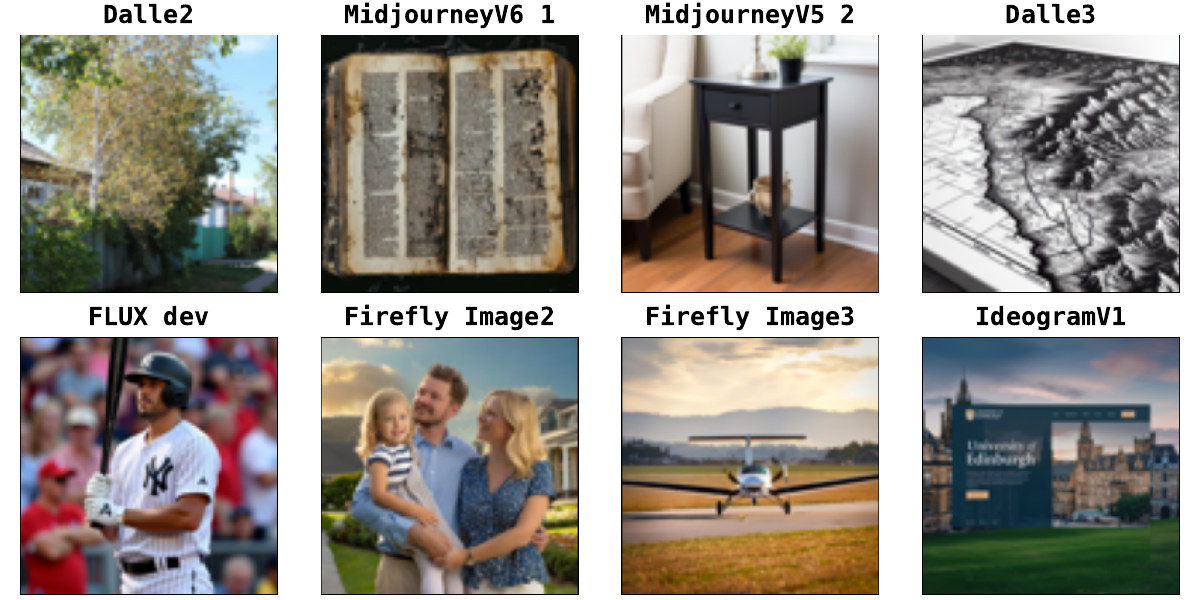}
            \caption{Commercial models (\commercialModels~models)}
            \label{fig:teaser_commercial}
            \vspace{0.4cm} %
        \end{subfigure}\vspace{-2mm}
    \caption{{\bf The {\em \datasetName{}} dataset.} Our dataset contains images sampled from three types of generative models. (a) We systematically download open-source latent diffusion models from a model-sharing community~\cite{von-platen-etal-2022-diffusers,HuggingFaceDiffusersWeb}. (b) We select popular open source generators with a variety of architectures and training procedures. (c) We sample from both closed and open state-of-the-art commercial models. We present example images and their corresponding model names.}
    \vspace{-4.4mm}
    \label{fig:teaser}
\end{figure*}

We hypothesize that the lack of diversity in training datasets is a major source of these shortcomings. 
Although today's datasets often contain millions of fake images, they come from a relatively small number of generators.
As a result, this data fails to capture many sources of variation that one might encounter in the wild. These limitations also make it challenging to accurately benchmark performance, since it is easy for cues that work well on one set of generators to fail on others.

To address these problems, we propose {\em Community Forensics}, a dataset that is significantly more diverse and comprehensive than those in prior works (\cref{fig:teaser}).
Our dataset contains images generated by: (a) thousands of systematically downloaded open-source latent diffusion models, (b) hand-selected open source models with various architectures, and (c) state-of-the-art commercial models.

To acquire large numbers of models, we sample images from thousands of text-to-image diffusion models hosted on a popular model-sharing website, Hugging Face~\cite{HuggingFaceDiffusersWeb}. 
We exploit the fact that these models use a common programming library~\cite{von-platen-etal-2022-diffusers} and thus can be sampled in a standardized way. A large fraction of them are extensions of Stable Diffusion~\cite{rombach2022LDM}, but collectively capture a variety of common model variations, such as in the architecture, image processing, and image content.
We also sample images from many other open source models, including GANs~\cite{goodfellow2014generative}, autoregressive models~\cite{gregor2014deep}, and consistency models~\cite{song2023consistency,luo2023lcmLora}. 
To ensure sufficient diversity in the image content of the generated images, we sample generators using captions sourced from a mix of several existing datasets whenever possible.

Our dataset contains \modelNum{} distinct models, roughly $\modelSizeMult{}\times$ more than the most extensive previous dataset that sample images from generative models~\cite{Wang_2020_CVPR,Ojha_2023_CVPR,epstein2023online,cozzolino2024raising,bammey2023synthbuster,zhu2024genimage,guo2024tracing,asnani2023reverse},
and covers a variety of recent model designs (\cref{fig:teaser}). %

We use this dataset to study generalization in the generated image detection problem.
Our experiments confirm that increasing the diversity of data improves generalization, consistent with previous findings in other areas of computer vision~\cite{schuhmann2021laion,schuhmann2022laion5b,hendrycks2021robustOODdata}. In contrast to other work, our dataset aims to increase diversity via \textit{number of generators}, 
an underexplored axis of diversity.
Through experiments, we find:
\begin{itemize}[leftmargin=*,topsep=1pt]
\item {\em Classifiers trained on our dataset obtain strong performance}, both on our newly proposed evaluations and on multiple previously-proposed benchmarks.
\item {\em Adding more generative models improves generatlization}. 
\cref{fig:nummodels} demonstrates the performance of fake image detection when trained on samples from varying numbers of diffusion models. 
Notably, the performance improves as more models are added, even across different architectures.
\item {\em Including diverse generative model architectures significantly improves results}, since classifiers do not fully generalize between generator architectures. 
\item {\em Standard classifiers perform well.} In contrast to observations from recent work, we find that end-to-end training of classifiers based on CNNs or ViTs generalizes well, with qualitatively similar to that of other recognition problems. 
\end{itemize}

\secVspace{}
\vspace{-0.5mm}
\section{Related work}
\label{sec:related}
\vspace{2mm}
\paragraph{Datasets for detecting generated images.} 
A number of datasets have been proposed for specifically detecting ``deepfake'' images containing manipulated faces~\cite{rossler2018faceforensics,rossler2019faceforensics++,dolhansky2020deepfake,kwon2021kodf,li2020celeb,khalid2021fakeavceleb,zi2020wilddeepfake}.
Rather than focusing on face manipulation, we address creating general-purpose methods that can detect images that have been directly produced by generative models.
Wang \etal{}~\cite{Wang_2020_CVPR} proposed a widely-used dataset of CNN-generated images, mixing images from GANs~\cite{karras2018proGAN,karras2019styleGAN,brock2018large,zhu2017cycleGAN,choi2018starGAN,park2019gauGAN} with other models~\cite{rossler2019faceforensics++,chen2017CRN,li2019imle,dai2019superRes,chen2018seeindark}.
This work showed that forensics models generalize between generative models, providing motivation for training on large datasets of diverse generators. However, their classifier was trained on images from a single GAN and was highly sensitive to data augmentation parameters, and more recent work shows that it does not generalize to newer models~\cite{corvi2023diffusionFingerprint,Ojha_2023_CVPR}. 
Ojha \etal{}~\cite{Ojha_2023_CVPR} introduced a dataset of recent diffusion models 
and found that training a linear classifier on CLIP features~\cite{radford2021clip} extracted from ProGAN-generated images performed well.
Cozzolino \etal{}~\cite{cozzolino2024raising} extend this work by studying the performance of CLIP-based detectors on various generative models and datasets.
Epstein \etal~\cite{epstein2023online} simulated detecting fake images in an online way by training a detector up to a certain year and testing it on generators released after that year. 
Zhu \etal~\cite{zhu2024genimage} collected 1.4M generated images from 8 different generators.
These datasets, however, only consider a handful of models (less than 20 each), limiting the generalization of their detectors. 
Asnani \etal proposed RED116~\cite{asnani2023reverse}, collecting 116K images from 116 generative models for predicting the model hyperparameters of a given generated image.
Guo \etal~\cite{guo2024tracing} extends RED116 to 140 generative models with 140K images total. 
We improve upon these works by collecting much more diverse generative models to improve the performance and generalization of the detector.
In concurrent work, Hong \etal~\cite{hong2024wildfake} acquires user-created images from Midjourney and CivitAI.
This strategy is complementary to ours: while it aims to collect in-the-wild fake images, its distribution is centered on images that users share, and the models are not necessarily identifiable, making it challenging to rigorously analyze the dataset's contents and to interpret experiments conducted on it.

\paragraph{Fingerprint-based image forensics methods.}
Classic work on image forensics relied on methods based on image statistics~\cite{popescu2005exposing} and physical constraints~\cite{johnson2007exposing}, rather than learning. A number of datasets have been created for detecting images that have been manipulated using traditional methods, such as with photo editors~\cite{dong2013casia,huh2018fighting,de2013exposing,Korus2016TIFS,ng2004data}. 
Recent works focus on detecting synthetic images by inspecting the generator fingerprints.
Zhang \etal{}~\cite{zhang2019GANartifacts} and Marra \etal{}~\cite{marra2019gansFingerprint} proposed identifying the spatial fingerprints left by the generator to detect synthetic images.
Others focus on spectral anomalies to detect synthetic images.
Durall \etal{}~\cite{durall2020cnnSpectral} and Dzanic \etal{}~\cite{dzanic2020CNNfourier} identified that CNN-generated images fail to reproduce certain spectral properties of real images.
Corvi \etal{}~\cite{corvi2023diffusionFingerprint} study the frequency fingerprints of the generated images and analyzes the cross-architecture generalization of the detector.
Bammey~\cite{bammey2023synthbuster} uses high-frequency artifacts to detect generated images. 
However, these approaches may be brittle since the artifacts they rely on can be eliminated by post-processing~\cite{cozzolino2024raising}. 
We instead approach this problem in a data-driven manner, scaling the number of models, images, and architectures. 
Recent work has created ensembles of fake image classifiers~\cite{hou2024deepfake}.
In parallel, researchers have detected text generated by language models using supervised learning and heuristics~\cite{mitchell2023detectgpt,jawahar2020automatic,bakhtin2019real,uchendu2020authorship,gehrmann2019gltr,sadasivan2023can,krishna2024paraphrasing,solaiman2019release}, which closely resemble those in visual forensics. However, no existing techniques that we are aware of aim to collect comprehensive datasets of community-created generators.

\paragraph{Out-of-distribution generalization.} 
Our work is related to the out-of-distribution recognition problem as it involves generalizing to unseen generators and image processing pipelines.
A variety of approaches have been proposed for this problem, based on likelihood ratios~\cite{ren2019likelihood,li2022likelihoodOOD,xiao2020likelihoodRegretOOD}, self-supervision~\cite{sehwag2020ssd,mohseni2020sslOOD,hendricks2019sslOOD,vyas2018sslOOD}, internal model statistics~\cite{hendrycks2016baseline,sastry2020oodGram}, temperature scaling~\cite{anirudh2023oodAnchoring,liang2018temperatureOOD}, and via energy-based models~\cite{du2021contrastiveEBM,liu2020energyOOD,elflein2021ebmOOD}.
Works by Schuhmann~\etal~\cite{schuhmann2022laion5b} and Hendrycks~\etal~\cite{hendrycks2021robustOODdata} show that diverse training data and data augmentation are important to improving the robustness to out-of-distribution samples. 
Our results are in line with these conclusions, as we find that a diverse set of generative models and stronger augmentations improve generalization.

\secVspace{}
\vspace{-3mm}
\section{The {\em \datasetName} Dataset} %
\secVspace{}
\label{sec:data_collection}

\begin{table*}[t]
\small
    \centering
    \begin{tabularx}{\textwidth}{Xllll} 
    \toprule
        Dataset & Models & Images & Architectures & Training setup \\
        \midrule
        Wang \etal{}~\cite{Wang_2020_CVPR} & 11 & 362K & GAN, Perceptual, Deepfake, ... & ProGAN~\cite{karras2018proGAN} \vs LSUN~\cite{yu15lsun} \\
        Ojha \etal{}~\cite{Ojha_2023_CVPR} & 4$^{*}$ & 10K$^{*}$ & GAN, Perceptual, Diffusion, ... & ProGAN~\cite{karras2018proGAN} \vs LSUN~\cite{yu15lsun} \\ 
        Epstein \etal{}~\cite{epstein2023online} & 14 & 570K & Diffusion & Diffusion \vs LAION~\cite{schuhmann2022laion5b}\\
        Cozzolino \etal{}~\cite{cozzolino2024raising} & 18 & 26K & Diffusion & LDM~\cite{rombach2022LDM} \vs MS-COCO~\cite{lin2014mscoco}\\
        Synthbuster~\cite{bammey2023synthbuster} & 9 & 10K & Diffusion & Diffusion \vs Dresden~\cite{gloe2010dresden}\\
        GenImage~\cite{zhu2024genimage} & 8 & 1.4M & Diffusion, GAN & Diffusion, GAN \vs ImageNet~\cite{deng2009imagenet}\\
        RED116~\cite{asnani2023reverse} & 116 & 116K & GAN, VAE, Autoregressive, ... & Many \vs{} Many\\
        RED140~\cite{guo2024tracing} & 140 & 140K & Diffusion, GAN, VAE, ... & Many \vs{} Many
        \vspace{0.3ex}\\\cdashline{1-5}[0.5pt/1pt] %
        \raisebox{0pt}[2.6ex][0ex]{Ours} & \totalModelsInclTest{} & \totalImages & Diffusion, GAN, Autoregressive, ... & Many \vs Many  \\
        \bottomrule
    \end{tabularx}
    \caption{{\bf Comparison with existing forensics datasets.} We compare the size of the dataset with existing datasets containing identifiable generative models. We only count the number of generated images. Our dataset contains significantly more generative models than prior works. $*$: Only counting the unique evaluation set by Ojha \etal{}~\cite{Ojha_2023_CVPR} as their dataset is based on Wang \etal{}~\cite{Wang_2020_CVPR}.} \figvspace{}
    \label{tab:other_datasets}
\end{table*}

To support our goal of studying generalization in generated image detection, we collect a dataset of images sampled from a wide range of models (Fig.~\ref{fig:teaser}). Our dataset consists of: (a) a large and systematically collected set of ``in-the-wild'' text-to-image latent diffusion models obtained from a model-sharing website, (b) hand-selected models from other open source architectures, and (c) closed and open state-of-the-art commercial models. 
We also pair these generated images with real images from other datasets. 
We preserve the original image format where possible, without any additional compression or resizing. This is to mitigate potential bias and performance degradation in out-of-distribution settings due to unwanted artifacts~\cite{grommelt2024fakeorjpeg,guillaro2024biasfreetrainingparadigmgeneral,rajan2025aligneddatasetsimprovedetection}. 
Our dataset contains significantly more models than previous works (\cref{tab:other_datasets}) and spans a wider range of architectures, processing pipelines, and semantic content.

\vspace{-1mm}
\subsection{Systematically collecting generative models}
\subsecVspace{}
\label{sec:automated_collection}
We perform our systematic collection using publicly available, open source\footnote{We use ``open source'' to describe models with public weights and code, even if they may be closed in some aspect (\eg, private training data).} 
models that use the Hugging Face {\tt diffusers} library~\cite{von-platen-etal-2022-diffusers,HuggingFaceDiffusersWeb} because: 1) it is a popular library for creating text-to-image models and is widely used by hobbyists, 2) thousands of such models are publicly indexed, and 3) it provides a standard interface by which we can sample images.
We process the models in the order of popularity, as indicated by the number of downloads. 
Our pipeline downloads each model and extracts relevant hyperparameters (e.g., number of diffusion steps and guidance scale), sampling pipeline configurations, and metadata, from both the model cards on the model-sharing webpage and the metadata retrieved by the \texttt{diffusers} library~\cite{von-platen-etal-2022-diffusers,HuggingFaceDiffusersWeb}.
We sample images using a distribution of text prompts obtained from various real datasets (\cref{sec:realimages}). 
Since experiments suggest diminishing returns for repeatedly sampling from any given model, we sample a few hundred images from each one. 
We sample \hfModelNum{} models with approximately \hfImagesAvg{} images each, for a total of \hfImagesTotal{} images from this process.

While the lack of documentation in each model and the scale of data collection make it challenging to exactly characterize the model designs in this set, they appear to be entirely (or almost entirely) based on latent diffusion. 
More specifically, we categorize models as being based on {\em latent diffusion} if they perform a denoising process on a latent representation.\footnote{We note that this definition includes latent consistency models~\cite{luo2023latent}.} %
Based on this criterion and the self-reported tags, all models in our systematically collected set appear to be based on latent diffusion. 
While pixel-based diffusion models also use the {\tt diffusers} library (e.g., DeepFloyd~\cite{DeepFloyd2024}), they were incompatible with our automated generation pipeline.
We record such incompatible models and manually sample them to either construct an out-of-distribution test set~(\cref{sec:evalset}), or as manually-chosen models for training data~(\cref{sec:manual_models}). %

We show examples of the sampled images in Fig.~\ref{fig:teaser}. 
\supparxiv{In the supplement}{In \Cref{appendix:example_model_card}}, we provide examples of models and information from their project pages. 
These models generate a variety of images, with various types of semantic content and preprocessing. %
For example, a large fraction of these models adapt variations of a popular pretrained latent diffusion model, Stable Diffusion~\cite{rombach2022LDM}, to different downstream applications, and use a number of adaptation strategies (e.g., using LoRA~\cite{hu2021lora}). 
We provide the model metadata with each image to enable other possible forensics applications. 
We discuss these \supparxiv{in the supplement}{in \cref{appendix:model_metadata}} and provide information about image and model licenses.

\subsecVspace{}
\subsection{Collecting images from other architectures} \subsecVspace{}
\vspace{9pt}
\paragraph{Images from manually chosen models.}
\label{sec:manual_models}
To ensure that our dataset contains a broader range of models, we manually select \manualModelNum{} models from public repositories and sample an average of \manualImagesAvg{} images per model, resulting in \manualImagesTotal{} images total. We note that this number is itself on par with (or more than) many prior datasets with identifiable generative models~\cite{Wang_2020_CVPR,Ojha_2023_CVPR,epstein2023online,cozzolino2024raising,bammey2023synthbuster,zhu2024genimage}. We include several GANs (e.g., StyleGANs~\cite{karras2020stylegan2,karras2020stylegan2ada,karras2021stylegan3,sauer2022styleganxl}, BigGAN~\cite{brock2018large}, StyleSwin~\cite{zhang2022styleswin}, GigaGAN~\cite{kang2023gigagan}, ProGAN~\cite{karras2018proGAN}, ProjectedGAN~\cite{Sauer2021projectedGAN}, GANsformer~\cite{hudson2021gansformer}, SAN~\cite{takida2024san}, and CIPS~\cite{anokhin2020cips}), pixel-based diffusion models (e.g., GLIDE~\cite{nichol2022glide}, ADM~\cite{dhariwal2021guided}, and DeepFloyd~\cite{DeepFloyd2024}), latent diffusion models (e.g., VQ-Diffusion~\cite{gu2022vqdiff}, Diffusion Transformers~\cite{peebles2023dit}, and Latent Flow Matching~\cite{dao2023lfm}), and an autoregressive model (Taming Transformers~\cite{esser2021taming}).

\paragraph{Images from commercial models.}
\label{sec:commercial_models}
We sample \commercialImages{} images from \commercialModels{} commercial models using LAION-based captions to evaluate the generalization to state-of-the-art models with typically unknown architectures: DALL·E 2, 3~\cite{dalle2,dalle3}, Ideogram V1, V2~\cite{ideogram}, Midjourney V5, V6~\cite{midjourney2022}, Firefly Image 2, 3~\cite{firefly2023}, FLUX.1-dev, schnell~\cite{flux}, and Imagen 3~\cite{imagen3}. 

\subsecVspace{}
\subsection{Collecting real images} \subsecVspace{}
\label{sec:realimages}
To help study how real images influence forensics models, we source real images from a variety of existing datasets: LAION~\cite{schuhmann2021laion}, ImageNet~\cite{deng2009imagenet}, COCO~\cite{lin2014mscoco}, FFHQ~\cite{karras2019styleGAN}, CelebA~\cite{liu2015celeba}, MetFaces~\cite{karras2020stylegan2ada}, AFHQ~\cite{choi2020afhqv2}, Forchheim~\cite{hadwiger2021forchheim}, IMD2020~\cite{novozamsky2020imd2020}, Landscapes HQ~\cite{Skorokhodov2021landscapes}, and VISION~\cite{shullani2017vision}.\footnote{Following common convention, we refer to these images as {\em real} images, even though they may be synthetic (e.g., graphic design). More precisely, our goal is to distinguish ``AI-generated'' images from the originals. }

\subsecVspace{}
\subsection{Curating the evaluation set} \subsecVspace{}
\label{sec:evalset}
We construct our evaluation set using the incompatible models from our automated sampling pipeline, commercial models (\cref{sec:commercial_models}), and manually collected open source models.
The evaluation set comprises \evalImagesUnseen{} images sampled from \evalModelsUnseen{} models not included in the training set. This includes our commercial models set and an additional \evalImagesOOD{} images from \evalModelsOOD{} models: 
Deci Diffusion V2~\cite{DeciFoundationModels}, GALIP~\cite{tao2023galip}, KandinskyV2.2~\cite{kandinsky2023models}, Kvikontent~\cite{KvikontentMidjourney}, LCM-LoRA-SDv1.5, LCM-LoRA-SDXL, LCM-LoRA-SSD1B~\cite{luo2023lcmLora}, Stable Cascade~\cite{pernias2023stablecascade}, DF-GAN~\cite{tao2022df}, and HDiT~\cite{crowson2024scalable}, sampled using RAISE~\cite{dang2015raise}, ImageNet~\cite{deng2009imagenet}, FFHQ~\cite{karras2019styleGAN}, and COCO~\cite{lin2014mscoco}-based captions.

The generated images are paired with the source real data that are used to prompt the generators.
However, since some of the real datasets do not have appropriate licenses for redistribution (e.g., LAION~\cite{schuhmann2021laion,schuhmann2022laion5b}), we created a \textit{public} version of our evaluation set by pairing the generated images with openly licensed COCO~\cite{lin2014mscoco} and FFHQ~\cite{karras2019styleGAN} which allow redistribution for non-commercial purposes.
The \textit{public} version of our evaluation set will serve as an easily reproducible and shareable evaluation set that will complement our default set. We will refer to our default set as the \textit{comprehensive} evaluation set.
We \supparxiv{will}{} also release the instructions to reconstruct our \textit{comprehensive} set. However, note that it may not be possible to exactly reconstruct this set in the future due to link rot.

\subsecVspace{}
\subsection{Generating images}
\subsecVspace{}
\label{sec:generating_images}
Unconditional models are sampled until we reach the desired number of images.
For class conditional models, we sample an equal number of images per class.
Text-conditional models are sampled using captions obtained from real images (Sec.~\ref{sec:realimages}).  
We either use captions that are already present in the dataset or use BLIP~\cite{li2022blip} to generate them.
Some models such as GigaGAN~\cite{kang2023gigagan} and HDiT~\cite{crowson2024scalable} do not provide a pretrained model, so we instead use their pre-generated images.
Generated images are saved in PNG format to avoid compression artifacts.
However, Firefly~\cite{firefly2023} generated images are saved in JPEG format as their web UI 
only allows downloading in JPEG.

\vspace{-2mm}
\section{Experiments}
\secVspace{}
\label{sec:experiments}

\definecolor{high}{HTML}{98df8a}  %
\definecolor{mid}{HTML}{ffbb78}  %
\definecolor{low}{HTML}{ec462e}%
\newcommand*{\opacity}{90}%
\newcommand*{\minval}{0.5}%
\newcommand*{\midval}{0.8}%
\newcommand*{\maxval}{1.0}%
\newcommand{\gradient}[1]{
    \ifdimcomp{#1pt}{>}{\maxval pt}{\cellcolor{high!\opacity} #1}{
        \ifdimcomp{#1pt}{<}{\minval pt}{\cellcolor{low!\opacity} #1}{
            \ifdimcomp{#1pt}{>}{\midval pt}{
                \pgfmathparse{int(round(100*(#1/(\maxval-\midval))-(\midval*(100/(\maxval-\midval)))))}
                \xdef\tempa{\pgfmathresult}
                \cellcolor{high!\tempa!mid!\opacity} #1
            }{
                \pgfmathparse{int(round(100*(#1/(\midval-\minval))-(\minval*(100/(\midval-\minval)))))}
                \xdef\tempa{\pgfmathresult}
                \cellcolor{mid!\tempa!low!\opacity} #1
            }
        }
    }
}
\newcommand{\gradientbf}[1]{
    \ifdimcomp{#1pt}{>}{\maxval pt}{\cellcolor{high!\opacity} \textbf{#1}}{
        \ifdimcomp{#1pt}{<}{\minval pt}{\cellcolor{low!\opacity} \textbf{#1}}{
            \ifdimcomp{#1pt}{>}{\midval pt}{
                \pgfmathparse{int(round(100*(#1/(\maxval-\midval))-(\midval*(100/(\maxval-\midval)))))}
                \xdef\tempa{\pgfmathresult}
                \cellcolor{high!\tempa!mid!\opacity} \textbf{#1}
            }{
                \pgfmathparse{int(round(100*(#1/(\midval-\minval))-(\minval*(100/(\midval-\minval)))))}
                \xdef\tempa{\pgfmathresult}
                \cellcolor{mid!\tempa!low!\opacity} \textbf{#1}
            }
        }
    }
}
\newcommand{\gradientcell}[6]{
    \ifdimcomp{#1pt}{>}{#3 pt}{#1}{
        \ifdimcomp{#1pt}{<}{#2 pt}{#1}{
            \pgfmathparse{int(round(100*(#1/(#3-#2))-(\minval*(100/(#3-#2)))))}
            \xdef\tempa{\pgfmathresult}
            \cellcolor{#5!\tempa!#4!#6} #1
    }}
}

\newcommand{\rowheight}{2.2}

\begin{table*}[t!]
    \def\arraystretch{1.3}
    \setlength{\tabcolsep}{1pt}
    \small
        \centering
        \begin{adjustbox}{max width=\textwidth} %
        \setlength{\aboverulesep}{0pt}
        \setlength{\belowrulesep}{0pt}
        \dashgapcolor{lightgray}
        \begin{tabular}{lcccccc;{1pt/1pt}c|cccccc;{1pt/1pt}c}\toprule
            \multirow{3.3}{*}{Model} & \multicolumn{7}{c|}{Evaluation Set (mAP)} & \multicolumn{7}{c}{Evaluation Set (Acc)}\\
            \cmidrule(){2-8}\cmidrule(){9-15} %
            & \multirow{\rowheight}{*}{\makecell[c]{Wang \etal{}\\\cite{Wang_2020_CVPR}}} & \multirow{\rowheight}{*}{\makecell[c]{Ojha \etal{}\\\cite{Ojha_2023_CVPR}}} & \multirow{\rowheight}{*}{\makecell[c]{SB\\\cite{bammey2023synthbuster}}} &\multirow{\rowheight}{*}{\makecell[c]{GenImage\\\cite{zhu2024genimage}}} & \multicolumn{2}{c;{1pt/1pt}}{Ours} & \multirow{\rowheight}{*}{Mean} & \multirow{\rowheight}{*}{\makecell[c]{Wang \etal{}\\\cite{Wang_2020_CVPR}}} & \multirow{\rowheight}{*}{\makecell[c]{Ojha \etal{}\\\cite{Ojha_2023_CVPR}}} & \multirow{\rowheight}{*}{\makecell[c]{SB\\\cite{bammey2023synthbuster}}} &\multirow{\rowheight}{*}{\makecell[c]{GenImage\\\cite{zhu2024genimage}}} & \multicolumn{2}{c;{1pt/1pt}}{Ours} & \multirow{\rowheight}{*}{Mean}\\
            \cmidrule(lr){6-7}\cmidrule(lr){13-14}
            & & & & & Comp. & Public & & & & & & Comp. & Public & \\\midrule
            Wang \etal{}~\cite{Wang_2020_CVPR} & \gradient{0.897} & \gradient{0.696} & \gradient{0.516} & \gradient{0.642} & \gradient{0.537} & \gradient{0.600} & \gradient{0.648} & \gradient{0.714} & \gradient{0.527} & \gradient{0.508} & \gradient{0.533} & \gradient{0.513} & \gradient{0.517} & \gradient{0.552} \\
            Ojha \etal{}~\cite{Ojha_2023_CVPR} & \gradient{0.939} & \gradient{0.957} & \gradient{0.620} & \gradient{0.797} & \gradient{0.592} & \gradient{0.656} & \gradient{0.760} & \gradient{0.791} & \gradient{0.821} & \gradient{0.532} & \gradient{0.641} & \gradient{0.540} & \gradient{0.548} & \gradient{0.646} \\
            GenImage~\cite{zhu2024genimage} & \gradient{0.929} & \gradient{0.984} & \gradient{0.813} & \gradientbf{0.999} & \gradient{0.912} & \gradient{0.968} & \gradient{0.934} & \gradient{0.795} & \gradient{0.966} & \gradient{0.719} & \gradientbf{0.990} & \gradient{0.818} & \gradient{0.886} & \gradient{0.862} \\
            $^{\blacklozenge}$RED140~\cite{guo2024tracing} -\textit{ High res.} & \gradient{0.900} & \gradient{0.954} & \gradient{0.765} & \gradient{0.927} & \gradient{0.764} & \gradient{0.861} & \gradient{0.862} & \gradient{0.694} & \gradient{0.780} & \gradient{0.558} & \gradient{0.674} & \gradient{0.562} & \gradient{0.565} & \gradient{0.639} \\\cmidrule(){1-15}
            Ours & \gradient{0.964} & \gradient{0.991} & \gradient{0.904} & \gradient{0.990} & \gradient{0.971} & \gradient{0.977} & \gradient{0.966} & \gradient{0.873} & \gradient{0.950} & \gradient{0.818} & \gradient{0.946} & \gradient{0.861} & \gradient{0.888} & \gradient{0.889} \\
            Ours - \textit{High res.} & \gradientbf{0.967} & \gradientbf{0.996} & \gradientbf{0.974} & \gradient{0.998} & \gradientbf{0.987} & \gradientbf{0.994} & \gradientbf{0.986} & \gradientbf{0.901} & \gradientbf{0.970} & \gradientbf{0.908} & \gradient{0.957} & \gradientbf{0.892} & \gradientbf{0.912} & \gradientbf{0.923} \\\bottomrule
        \end{tabular}
        \end{adjustbox}
        \vspace{-2mm}
        \caption{{\bf Generalization of AI-generated image detectors across benchmarks.} We evaluate the classifiers trained on our dataset on several benchmarks, including our own. 
        We also evaluate several previously released classifiers. 
        Our \textit{Comprehensive} set (abbreviated as \textit{Comp.}) pairs the generated images with original real data; the \textit{Public} set pairs them with openly licensed COCO~\cite{lin2014mscoco} and FFHQ~\cite{karras2019styleGAN} for license-compliant redistribution of the evaluation set (\cref{sec:evalset}).
        We use plain CLIP-ViT-S~\cite{dosovitskiy2020vit,Ilharco2021OpenCLIP,radford2021clip} architecture with $224^2$ and $384^2$ (\textit{High res.}) input resolutions, Wang \etal{}~\cite{Wang_2020_CVPR} and GenImage~\cite{zhu2024genimage} use ResNet-50~\cite{he2016resnet} with $224^2$ input resolution, and Ojha \etal{}~\cite{Ojha_2023_CVPR} uses CLIP-ViT-L with $224^2$ input resolution as the backbone. 
        Our classifiers show robust performance across all evaluation sets, outperforming all baselines in out-of-distribution settings (\cite{Wang_2020_CVPR,Ojha_2023_CVPR,bammey2023synthbuster} and \textit{Ours}) and nearly matches GenImage~\cite{zhu2024genimage} on its {\em in-distribution} evaluation set.
        $\blacklozenge$: RED140~\cite{guo2024tracing} is trained following our training procedure with $384^2$ input resolution as they do not provide a trained classifier. 
        }\figvspace
        \label{tab:comparison_priorWorks}
\end{table*}

We use our dataset to conduct a study of generalization in visual forensics, asking a number of questions: {\bf (1)} How well do forensics models trained on our dataset generalize to unseen models? {\bf (2)} Does adding more models improve detection performance? {\bf (3)} How does diversity of the training data affect performance? {\bf (4)} What architectures and data augmentation schemes are most successful?

\subsecVspace{}
\subsection{Training image forensics models}
\subsecVspace{}

We train binary classifiers that detect generated images using our dataset to study the generalization in image forensics.
We construct our training set of 5.4M images by pairing \totalImages{} generated images with \totalImages{} real images. 

\paragraph{Training and evaluation setup.} We evaluate the models trained on our dataset and compare them with prior works~\cite{Ojha_2023_CVPR,Wang_2020_CVPR,zhu2024genimage,guo2024tracing}.
As RED140~\cite{guo2024tracing} already contains RED116~\cite{asnani2023reverse}, we do not compare with RED116 dataset.
Following prior works~\cite{Ojha_2023_CVPR,Wang_2020_CVPR}, we use the threshold-independent mean average precision (mAP) and accuracy (Acc.) as our evaluation metrics.
We compute the mAP and accuracy by averaging the results of each generative model.
We use five evaluation sets: Wang \etal{}~\cite{Wang_2020_CVPR}, Ojha \etal{}~\cite{Ojha_2023_CVPR}, Synthbuster~\cite{bammey2023synthbuster}, GenImage~\cite{zhu2024genimage}, and our evaluation set.
All evaluation sets apart from GenImage~\cite{zhu2024genimage} evaluate out-of-distribution performance for all classifiers. 
GenImage~\cite{zhu2024genimage} evaluation set, however, contains the same set of generators used in training, and is an in-distribution evaluation set for their classifiers.
Concretely, the evaluation set by Wang \etal{}~\cite{Wang_2020_CVPR} and Ojha \etal{}~\cite{Wang_2020_CVPR,Ojha_2023_CVPR} contains models such as CRN~\cite{chen2017CRN}, CycleGAN~\cite{zhu2017cycleGAN}, DALL$\cdot$E~\cite{dalle2}, DeepFake~\cite{dolhansky2020deepfake},  IMLE~\cite{li2019imle}, SAN~\cite{dai2019superRes}, StarGAN~\cite{choi2018starGAN}, and SITD~\cite{chen2018seeindark} which are unseen by both their and our classifiers.
Synthbuster~\cite{bammey2023synthbuster} evaluation set is comprised of RAISE~\cite{dang2015raise}-based synthetic images of DALL$\cdot$E~\cite{dalle2,dalle3}, Firefly~\cite{firefly2023}, Glide~\cite{nichol2022glide}, Midjourney~\cite{midjourney2022}, and Stable Diffusion~\cite{rombach2022LDM,podell2023sdxl}, and is mostly out of distribution for all classifiers.
GenImage~\cite{zhu2024genimage} evaluation set is a validation split of their training set; they use an identical set of models to train their classifier: ADM~\cite{dhariwal2021guided}, BigGAN~\cite{brock2018large}, Glide~\cite{nichol2022glide}, Midjourney~\cite{midjourney2022}, Stable Diffusion~\cite{rombach2022LDM}, VQ-Diffusion~\cite{gu2022vqdiff}, and Wukong~\cite{wukong2022}.

\paragraph{Model architecture.} 
Building on prior works which mainly used CLIP-ViT~\cite{Ilharco2021OpenCLIP,dosovitskiy2020vit,radford2021clip} and ResNet-50~\cite{he2016resnet}, we consider ViT~\cite{dosovitskiy2020vit} and ConvNeXt~\cite{liu2022convnext} pretrained models for our classifiers.
We use a plain ViT-S backbone~\cite{dosovitskiy2020vit} pretrained on CLIP objective~\cite{Ilharco2021OpenCLIP,radford2021clip} using LAION-2B~\cite{schuhmann2022laion5b}, ImageNet 21K, and ImageNet 1K datasets~\cite{deng2009imagenet}.
We also experiment with a ConvNeXt-S model~\cite{liu2022convnext} pretrained on ImageNet 21K and ImageNet 1K datasets~\cite{deng2009imagenet}.
We replace the classification head with a linear layer with sigmoid activation that outputs the probability of the image being generated.
Unlike prior works~\cite{cozzolino2024raising,Ojha_2023_CVPR} that freeze the CLIP-ViT backbone, we train the backbone end-to-end.
The models are obtained through \texttt{timm}~\cite{cherti2022reproducible,rw2019timm} library on Hugging Face. 
We experiment with two input resolutions, $224^2$ and $384^2$, to evaluate the impact of the input resolution on the detector's performance. 
We denote the detector with $384^2$ input resolution as \textit{High res.}
We implement the models using PyTorch~\cite{paszke2019pytorch}.
The hyperparameters are detailed \supparxiv{in the supplement.}{in \Cref{appendix:training_hyperparams}.}

\begin{figure}[t]
    \centering
    \includegraphics[width=\linewidth]{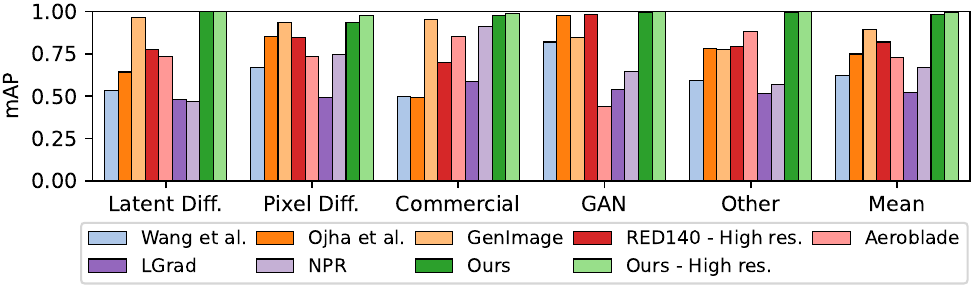}\vspace{-1mm}
    \caption{
    \textbf{Performance across generator types.} 
    We evaluate the classifier performance across five generator types -- latent diffusion, pixel diffusion, commercial models, GANs, and other architecture (Stable Cascade~\cite{pernias2023stablecascade}), and report the mean. We additionally evaluate three forensics methods, Aeroblade~\cite{ricker2024aeroblade}, LGrad~\cite{tan2023learning}, and NPR~\cite{tan2024rethinking}.
    Our classifiers show robust performance across all generator types, whereas prior works struggle to generalize.
    }\figvspace
    \label{fig:per_arch_eval}
\end{figure}

\paragraph{Data augmentation.} Prior work considered augmentations that were designed to simulate postprocessing, such as flipping, cropping, Gaussian blur, and JPEG recompression to train their detectors~\cite{Ojha_2023_CVPR,bammey2023synthbuster,cozzolino2024raising,Wang_2020_CVPR}. %
We propose an augmentation scheme that extends this approach and compare it with previously proposed augmentation methods. We expand the set of augmentations to handle additional transformations that can occur in the wild, such as padding, resizing, rotation, and shear, and integrate them into a framework that can apply complex sequences of transformations.
We introduce a modified version of RandAugment~\cite{cubuk2020randaugment} that applies a randomly-ordered sequence of augmentations to the images. 
Specifically, our modified RandAugment samples a random number $n$ between $0$ and $n_{\mathtt{max}}$ for each augmentation type.
Then, it applies the augmentations in random order until $n$ augmentations are applied for each augmentation type to the image.
We use various augmentations, including in-memory JPEG compression, random resizing with random interpolation methods, cropping, flipping, rotation, translation, shear, padding, and cutout. 

\begin{figure}[t]
    \centering
    \includegraphics[width=\linewidth]{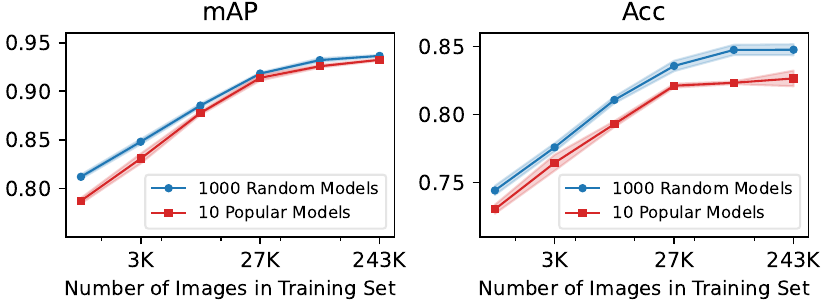}
    \vspace{-5mm}
    \caption{
    \textbf{Performance with increasing number of images.}
    We train a classifier with varying numbers of images from two sets: 1000 randomly chosen models and 10 popular (highly downloaded) models in the systematically collected subset. 
    We report the mean and standard error bands for each data point across 4 randomly sampled subsets.
    The classifier trained from 1000 random models outperforms 10 popular models in all cases. 
    Notably, the accuracy gap is wider than that of mAP, which may suggest that having a diverse set of models improves accuracy threshold calibration.
    }\figvspace %
    \label{fig:ablations_numImg}%
    
\end{figure}

\subsecVspace{}
\subsection{Generalization across benchmarks}
\subsecVspace{}
We first evaluate how well classifiers trained on our dataset perform across benchmarks. 
In \Cref{tab:comparison_priorWorks}, we observe that our models outperform the prior works~\cite{Ojha_2023_CVPR,Wang_2020_CVPR,zhu2024genimage,guo2024tracing} on all evaluation sets except GenImage. 
This is expected since the GenImage evaluation set is a validation split of their training set; all of the generators are already seen by their classifier.
On all other \textit{unseen} benchmarks, our classifiers outperform all prior works.
Notably, our classifiers achieve very high performance (0.987 mAP and 89.2\% accuracy) on our out-of-distribution, \textit{comprehensive} evaluation set, with a significant margin over prior works.
This gap in performance can be traced to our training data which incorporates a substantially richer variety of generators than existing works. 
Consequently, our classifiers demonstrate robust generalization to out-of-distribution data, where prior works often struggle.

To better illustrate the generalization of the classifiers, in \Cref{fig:per_arch_eval},
we split our \textit{comprehensive} evaluation set into five subsets and evaluate them: latent diffusion, pixel diffusion, commercial models, GANs, and other architecture type (Stable Cascade~\cite{pernias2023stablecascade}). 
Furthermore, we evaluate three additional forensics methods: 
Aeroblade~\cite{ricker2024aeroblade}, LGrad~\cite{tan2023learning}, and NPR~\cite{tan2024rethinking}. 
Our classifiers show strong performance across all generator types, unlike prior works which struggle to generalize.
For the following experiments, we use our best-performing model (\textit{High res.}) unless stated otherwise. %

\subsecVspace{}
\subsection{Impact of model diversity}
\subsecVspace{}
\label{subsec:modelnum_ab}

Next, we examine the impact of the number of models in training data. 
We train classifiers with images sampled from 3 to 3333 generators and evaluate them (\cref{fig:nummodels}).
To ensure that the gains are not due to simply sampling qualitatively different architectures, we only use our systematically collected latent diffusion models.
For each data point, we sample 10 random subsets of models with 100K training images each and report the mean and standard error bands.
We use an extended evaluation set that includes non-latent diffusion generators from our training set, which allows us to comprehensively assess the generalization capability of the classifiers trained exclusively on latent diffusion models.
We find that the performance steadily increases with the number of models. 
However, the performance begins to flatten out beyond 1000 models, suggesting diminishing returns. 
Interestingly, the performance also improves on out-of-distribution architectures such as GANs and pixel-based diffusion models, even though the classifier is only trained on latent diffusion models.

\begin{figure}[t]
    \centering
    \begin{subfigure}[t]{0.49\linewidth}
        \centering
        \raisebox{-\height}{\includegraphics[width=\linewidth]{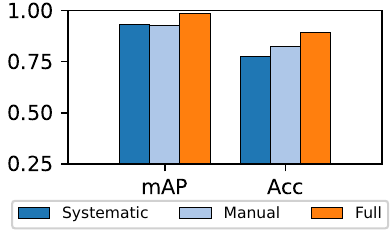}}
        \vspace{10.6pt}
        \caption{Impact of generator type diversity}
        \label{fig:ablations_generatorTypeSubset}
    \end{subfigure}
    \hfill
    \begin{subfigure}[t]{0.49\linewidth}
        \centering
        \raisebox{-\height}{\includegraphics[width=\linewidth]{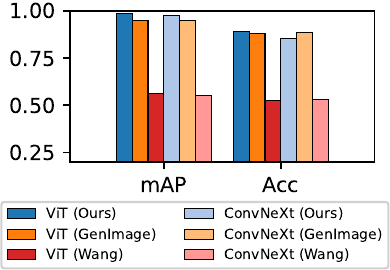}}
        \caption{
        Classifier backbone comparison
        }
        \label{fig:ablations_backbone}
    \end{subfigure}\vspace{-1mm}
    \caption{
    (a) {\bf Performance and model diversity.} We compare detection performance for commercial models using classifiers trained on different subsets of the dataset: the systematically collected latent diffusion models, the manually chosen models containing diverse generator types, and both. 
    As diversity increases, so does performance.
    (b) {\bf Classifier backbone comparison.} We compare the architectures across datasets: ours, GenImage~\cite{zhu2024genimage}, and Wang \etal{}~\cite{Wang_2020_CVPR}. Performance is similar between architectures.
    }\figvspace{}
    \label{fig:ablations_dataset_and_model}
\end{figure}

In \Cref{fig:ablations_numImg}, we vary the number of images from two sets: 1000 randomly chosen models and 10 popular models (as denoted by their number of downloads) downloaded from our systematically collected diffusion models.
While the results show that the performance improves with more training images, it begins to plateau at approximately 27K images. 
Moreover, the classifier trained on 1000 models outperforms the 10 models in all cases, indicating that model diversity is important for strong performance.
We also note that the accuracy gap is noticeably wider than that of mAP, which may suggest that model diversity is crucial in calibrating the accuracy thresholds of the classifiers.

Our experiments show that the performance improvements from increasing the number of models may plateau when they are limited to a single generator type (\cref{fig:nummodels}).
In \Cref{fig:ablations_generatorTypeSubset}, we show that the diversity of the generator type also plays a major role in generalization. 
We train classifiers on three different sets of training data: our systematically collected set, manually chosen set, and a full set consisting of both subsets.
The \textit{systematic} set comprises entirely of latent diffusion models, and the \textit{manual} set contains numerous generator types, including GANs, latent and pixel-based diffusion, and autoregressive models (\cref{sec:manual_models}).
The classifier trained on the \textit{manual} set with more diverse generator types shows similar performance compared to the one trained on the \textit{systematic} set, despite containing fewer than half the number of images (774K \vs{} 1.9M).
Additionally, we find that the two sets are complementary; the performance is further improved when we train using both sets.

\begin{figure}[t]
    \centering
    \begin{subfigure}{0.49\linewidth}
        \centering
        \includegraphics[width=\linewidth]{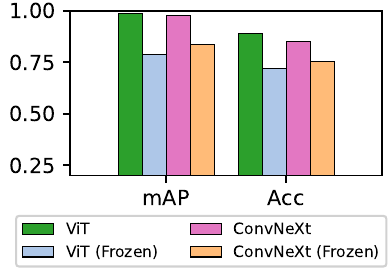}
        \caption{Impact of frozen backbones}
        \label{fig:ablations_frozenBB_a}
    \end{subfigure}
    \hfill
    \begin{subfigure}{0.49\linewidth} 
        \centering 
        \includegraphics[width=\linewidth]{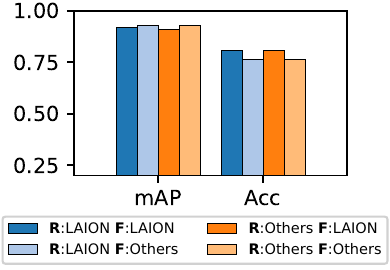}
        \caption{Semantic alignment analysis}
        \label{fig:ablations_semanticAlignment}
    \end{subfigure}\vspace{-1.0mm}
    \caption{%
    (a) \textbf{Evaluating frozen backbones.} Freezing the pretrained backbone, a common practice in prior works~\cite{Ojha_2023_CVPR,cozzolino2024raising}, consistently decreases the performance. (b) \textbf{Analyzing source and generated data alignment.} We evaluate how the pairing of the real datasets affects performance. `\textbf{R}' denotes the real dataset used in training, and `\textbf{F}' indicates the source dataset used to obtain the captions for prompting the generators. The results suggest that pairing the source data (i.e., real data used to prompt the generators) with the generated images is not essential for performance.}
    \label{fig:ablations_frozen_alignment}\figvspace
\end{figure}

\subsecVspace{}
\subsection{Analysis of design choices}
\subsecVspace{}

We examine the impact of various design choices, including some suggested in earlier works.
In particular, we investigate the choice of backbone models, freezing the backbone, semantic alignment between the real and generated data, and robustness to transformations. %

\paragraph{Classifier backbone comparison.} We compare the performance of the classifier trained using CLIP-ViT~\cite{dosovitskiy2020vit,Ilharco2021OpenCLIP,radford2021clip} and ConvNeXt~\cite{liu2022convnext} backbones following our training procedure in \Cref{fig:ablations_backbone}. 
We examine three datasets: ours, GenImage~\cite{zhu2024genimage}, and Wang \etal{}~\cite{Wang_2020_CVPR}.
We observe similar performance between architectures across all datasets.

\paragraph{Frozen backbone.}
Prior works~\cite{Ojha_2023_CVPR,cozzolino2024raising} suggested using a frozen CLIP-ViT backbone for training the classifiers.
We study this by training the classifiers with both frozen and unfrozen pretrained backbones, using CLIP-ViT~\cite{dosovitskiy2020vit,radford2021clip,Ilharco2021OpenCLIP} and ConvNeXt~\cite{liu2022convnext}.
As shown in Fig.~\ref{fig:ablations_frozenBB_a}, freezing the backbone consistently leads to worse results, indicating that 
end-to-end training is crucial for high performance.

\paragraph{Semantic alignment.}
Existing works often pair the generated images with the source dataset (i.e., the real dataset used to prompt or generate the images) arguing that misaligned data can introduce  bias~\cite{Wang_2020_CVPR,Ojha_2023_CVPR,cozzolino2024raising,bammey2023synthbuster}.
We test this practice in \cref{fig:ablations_semanticAlignment} by 
examining the performance with both semantically aligned and misaligned real datasets.
Specifically, we consider two real datasets: one comprised exclusively from LAION~\cite{schuhmann2021laion} and another combining ImageNet~\cite{deng2009imagenet}, MS-COCO~\cite{lin2014mscoco}, LandscapesHQ~\cite{Skorokhodov2021landscapes}, Forchheim~\cite{hadwiger2021forchheim}, VISION~\cite{shullani2017vision}, and IMD2020~\cite{novozamsky2020imd2020}.
We sample our systematically collected latent diffusion models using these two sets and categorize the generated images by their source dataset.
The performance differences of all pairs were marginal, suggesting that strict alignment may not be as critical with sufficiently large data. %

\begin{figure}[t]
    \centering
    \includegraphics[width=\linewidth]{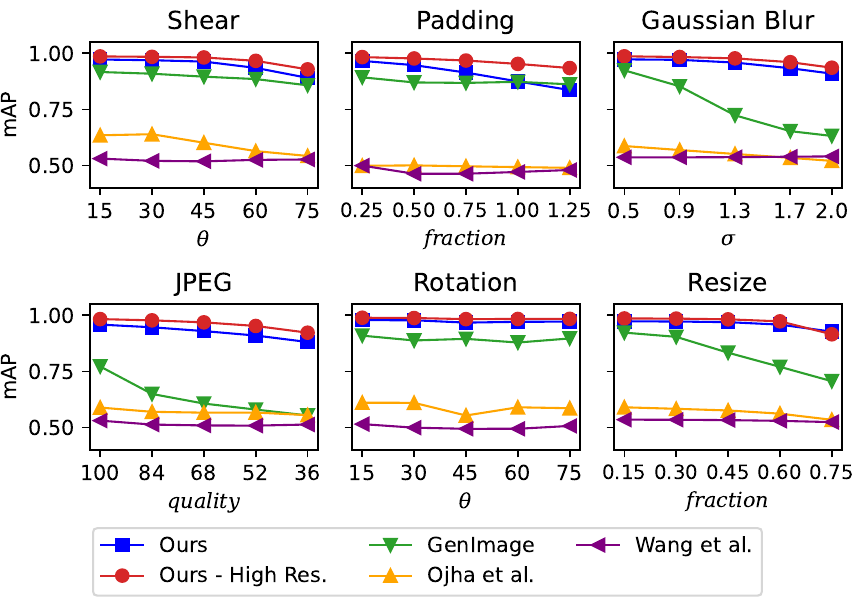}
    \vspace{-6mm}
    \caption{\textbf{Robustness to various transformations.} Our classifiers display robust performance across transformations. Other works generally show more sensitivity to these factors.}
    \label{fig:ablations_augmentation_strength}\figvspace{}
\end{figure}

\paragraph{Robustness to transformations.}
\Cref{fig:ablations_augmentation_strength} illustrates the robustness of the classifiers against various transformations.
Following prior works~\cite{Ojha_2023_CVPR,Wang_2020_CVPR}, we test robustness to JPEG compression and Gaussian blur.
Additionally, we examine robustness to rotation, resizing, padding, and shear, as they commonly occur in real-world scenarios. 
For \textit{padding}, we randomly pad the width or height of the image with a given fraction and scale it back to the original size.
Similarly, in \textit{resize}, we randomly upsample or downsample the height and width of the image by a given fraction and resize it back to the original size (e.g., if \textit{fraction} is $0.3$, resize the height and width to $0.7\times$ or $1.3\times$ and then scale it back to the original size).
The results 
demonstrate that our models are more robust to transformations than existing models.
Specifically, GenImage~\cite{zhu2024genimage} is notably more sensitive to Gaussian blur, JPEG compression, and resizing artifacts;
classifier by Ojha \etal{}~\cite{Ojha_2023_CVPR} displays sensitivity to \textit{shear} transforms, and the one by Wang \etal{}~\cite{Wang_2020_CVPR} performs poorly overall.

\subsecVspace{}
\subsection{Other forensics applications}
\subsecVspace{}

Our dataset may enable further forensics studies that can take advantage of our diverse array of generators.
To illustrate this, \supparxiv{in the supplement}{in \cref{appendix:other_applications}},
we identify the type of generator used to synthesize a given image
by using $k$-nearest neighbor in the feature space of our classifier.

\section{Conclusion}
\secVspace{}
\label{sec:discussion}

In this paper, we studied the problem of generalizing to unseen generative models in synthetic image detection.
We proposed a new dataset, \textit{\datasetName{}}, which contains \modelNum{} models and \totalImages{} images collected from various public sources.
We studied the impact of model diversity and demonstrated that it plays a crucial role in enhancing data diversity and generalization performance.
We trained classifiers on our dataset, studied their ability to generalize in various settings, and evaluated previously proposed models and training  practices.

We do not intend for our dataset to be used to train classifiers that are directly used in the wild. Detecting in-the-wild synthetic images remains a challenging open problem, and detection errors can have severe consequences (e.g., falsely accusing an author of creating fake images or allowing misinformation to be certified as real).
We hope that our work will serve as a stepping stone for future research in this area by providing tools and insights for studying generalization and data collection strategies.

\paragraph{Limitations.} While our dataset is diverse, a large portion of the data is diffusion-based, especially models based on Stable Diffusion~\cite{rombach2022LDM}.
Despite this, \Cref{fig:nummodels} demonstrates that adding more diffusion models to the training data improves the generalization capability of the classifiers across other generator types.
Furthermore, \Cref{fig:ablations_generatorTypeSubset} illustrates improved classifier performance when combining the diffusion-only \textit{systematic} set with a more diverse \textit{manual} set.
These results suggest that despite their similarities, each diffusion model may uniquely contribute through its semantic content, image processing pipelines, or architectural variations (\cref{sec:automated_collection}).
Future work may explore collecting more diverse models, including GANs, VQ-VAEs~\cite{van2017vqvae}, and autoregressive models.
We also note that the generative models sourced from the community may contain inappropriate content. 
While in many contexts it is important to detect such images, 
these generated images may require further scrutiny before being used in other downstream applications. 
Finally, although our experiments suggest that our forensics classifiers generalize to unseen models better than those of previous work, their error rates are still too high for many critical applications.

\paragraph{Acknowledgements.} We thank the creators of the many open source models that we used to collect the Community Forensics dataset. We thank Chenhao Zheng, Cameron Johnson, Matthias Kirchner, Daniel Geng, Ziyang Chen, Ayush Shrivastava, Yiming Dou, Chao Feng, Xuanchen Lu, Zihao Wei, Zixuan Pan, Inbum Park, Rohit Banerjee, and Ang Cao for the valuable discussions and feedback. This research was developed with funding from the Defense Advanced Research Projects Agency (DARPA) under Contract No. HR001120C0123.

{
    \small
    \bibliographystyle{ieeenat_fullname}
    \bibliography{main}

\begin{thebibliography}{138}
\providecommand{\natexlab}[1]{#1}
\providecommand{\url}[1]{\texttt{#1}}
\expandafter\ifx\csname urlstyle\endcsname\relax
  \providecommand{\doi}[1]{doi: #1}\else
  \providecommand{\doi}{doi: \begingroup \urlstyle{rm}\Url}\fi

\bibitem[Adobe(2023)]{firefly2023}
Adobe.
\newblock Firefly.
\newblock \url{https://www.adobe.com/products/firefly}, 2023.

\bibitem[AI(2024)]{ideogram}
Ideogram AI.
\newblock Ideogram.
\newblock \url{https://ideogram.ai}, 2024.

\bibitem[Anirudh and Thiagarajan(2023)]{anirudh2023oodAnchoring}
Rushil Anirudh and Jayaraman~J Thiagarajan.
\newblock Out of distribution detection via neural network anchoring.
\newblock In \emph{Asian Conference on Machine Learning}, pages 32--47. PMLR, 2023.

\bibitem[Anokhin et~al.(2020)Anokhin, Demochkin, Khakhulin, Sterkin, Lempitsky, and Korzhenkov]{anokhin2020cips}
Ivan Anokhin, Kirill Demochkin, Taras Khakhulin, Gleb Sterkin, Victor Lempitsky, and Denis Korzhenkov.
\newblock Image generators with conditionally-independent pixel synthesis.
\newblock \emph{arXiv preprint arXiv:2011.13775}, 2020.

\bibitem[Asnani et~al.(2023)Asnani, Yin, Hassner, and Liu]{asnani2023reverse}
Vishal Asnani, Xi Yin, Tal Hassner, and Xiaoming Liu.
\newblock Reverse engineering of generative models: Inferring model hyperparameters from generated images.
\newblock \emph{IEEE Transactions on Pattern Analysis and Machine Intelligence}, 45\penalty0 (12):\penalty0 15477--15493, 2023.

\bibitem[Bakhtin et~al.(2019)Bakhtin, Gross, Ott, Deng, Ranzato, and Szlam]{bakhtin2019real}
Anton Bakhtin, Sam Gross, Myle Ott, Yuntian Deng, Marc'Aurelio Ranzato, and Arthur Szlam.
\newblock Real or fake? learning to discriminate machine from human generated text.
\newblock \emph{arXiv preprint arXiv:1906.03351}, 2019.

\bibitem[Bammey(2023)]{bammey2023synthbuster}
Quentin Bammey.
\newblock Synthbuster: Towards detection of diffusion model generated images.
\newblock \emph{IEEE Open Journal of Signal Processing}, 2023.

\bibitem[Brock et~al.(2019)Brock, Donahue, and Simonyan]{brock2018large}
Andrew Brock, Jeff Donahue, and Karen Simonyan.
\newblock Large scale {GAN} training for high fidelity natural image synthesis.
\newblock In \emph{International Conference on Learning Representations}, 2019.

\bibitem[Chen et~al.(2018)Chen, Chen, Xu, and Koltun]{chen2018seeindark}
Chen Chen, Qifeng Chen, Jia Xu, and Vladlen Koltun.
\newblock Learning to see in the dark.
\newblock In \emph{Proceedings of the IEEE conference on computer vision and pattern recognition}, pages 3291--3300, 2018.

\bibitem[Chen and Koltun(2017)]{chen2017CRN}
Qifeng Chen and Vladlen Koltun.
\newblock Photographic image synthesis with cascaded refinement networks.
\newblock In \emph{Proceedings of the IEEE international conference on computer vision}, pages 1511--1520, 2017.

\bibitem[Cherti et~al.(2022)Cherti, Beaumont, Wightman, Wortsman, Ilharco, Gordon, Schuhmann, Schmidt, and Jitsev]{cherti2022reproducible}
Mehdi Cherti, Romain Beaumont, Ross Wightman, Mitchell Wortsman, Gabriel Ilharco, Cade Gordon, Christoph Schuhmann, Ludwig Schmidt, and Jenia Jitsev.
\newblock Reproducible scaling laws for contrastive language-image learning.
\newblock \emph{arXiv preprint arXiv:2212.07143}, 2022.

\bibitem[Choi et~al.(2018)Choi, Choi, Kim, Ha, Kim, and Choo]{choi2018starGAN}
Yunjey Choi, Minje Choi, Munyoung Kim, Jung-Woo Ha, Sunghun Kim, and Jaegul Choo.
\newblock Stargan: Unified generative adversarial networks for multi-domain image-to-image translation.
\newblock In \emph{Proceedings of the IEEE Conference on Computer Vision and Pattern Recognition}, 2018.

\bibitem[Choi et~al.(2020)Choi, Uh, Yoo, and Ha]{choi2020afhqv2}
Yunjey Choi, Youngjung Uh, Jaejun Yoo, and Jung-Woo Ha.
\newblock Stargan v2: Diverse image synthesis for multiple domains.
\newblock In \emph{Proceedings of the IEEE Conference on Computer Vision and Pattern Recognition}, 2020.

\bibitem[Corvi et~al.(2023)Corvi, Cozzolino, Zingarini, Poggi, Nagano, and Verdoliva]{corvi2023diffusionFingerprint}
Riccardo Corvi, Davide Cozzolino, Giada Zingarini, Giovanni Poggi, Koki Nagano, and Luisa Verdoliva.
\newblock On the detection of synthetic images generated by diffusion models.
\newblock In \emph{ICASSP 2023-2023 IEEE International Conference on Acoustics, Speech and Signal Processing (ICASSP)}, pages 1--5. IEEE, 2023.

\bibitem[Cozzolino et~al.(2024)Cozzolino, Poggi, Corvi, Nie{\ss}ner, and Verdoliva]{cozzolino2024raising}
Davide Cozzolino, Giovanni Poggi, Riccardo Corvi, Matthias Nie{\ss}ner, and Luisa Verdoliva.
\newblock Raising the bar of ai-generated image detection with clip.
\newblock In \emph{Proceedings of the IEEE/CVF Conference on Computer Vision and Pattern Recognition}, pages 4356--4366, 2024.

\bibitem[Crowson et~al.(2024)Crowson, Baumann, Birch, Abraham, Kaplan, and Shippole]{crowson2024scalable}
Katherine Crowson, Stefan~Andreas Baumann, Alex Birch, Tanishq~Mathew Abraham, Daniel~Z Kaplan, and Enrico Shippole.
\newblock Scalable high-resolution pixel-space image synthesis with hourglass diffusion transformers.
\newblock In \emph{Proceedings of the 41st International Conference on Machine Learning}, pages 9550--9575. PMLR, 2024.

\bibitem[Cubuk et~al.(2020)Cubuk, Zoph, Shlens, and Le]{cubuk2020randaugment}
Ekin~Dogus Cubuk, Barret Zoph, Jon Shlens, and Quoc Le.
\newblock Randaugment: Practical automated data augmentation with a reduced search space.
\newblock In \emph{Advances in Neural Information Processing Systems}, pages 18613--18624. Curran Associates, Inc., 2020.

\bibitem[Dai et~al.(2019)Dai, Cai, Zhang, Xia, and Zhang]{dai2019superRes}
Tao Dai, Jianrui Cai, Yongbing Zhang, Shu-Tao Xia, and Lei Zhang.
\newblock Second-order attention network for single image super-resolution.
\newblock In \emph{Proceedings of the IEEE/CVF conference on computer vision and pattern recognition}, pages 11065--11074, 2019.

\bibitem[Dang-Nguyen et~al.(2015)Dang-Nguyen, Pasquini, Conotter, and Boato]{dang2015raise}
Duc-Tien Dang-Nguyen, Cecilia Pasquini, Valentina Conotter, and Giulia Boato.
\newblock Raise: A raw images dataset for digital image forensics.
\newblock In \emph{Proceedings of the 6th ACM multimedia systems conference}, pages 219--224, 2015.

\bibitem[Dao et~al.(2023)Dao, Phung, Nguyen, and Tran]{dao2023lfm}
Quan Dao, Hao Phung, Binh Nguyen, and Anh Tran.
\newblock Flow matching in latent space.
\newblock \emph{arXiv preprint arXiv:2307.08698}, 2023.

\bibitem[De~Carvalho et~al.(2013)De~Carvalho, Riess, Angelopoulou, Pedrini, and de~Rezende~Rocha]{de2013exposing}
Tiago~Jos{\'e} De~Carvalho, Christian Riess, Elli Angelopoulou, Helio Pedrini, and Anderson de Rezende~Rocha.
\newblock Exposing digital image forgeries by illumination color classification.
\newblock \emph{IEEE Transactions on Information Forensics and Security}, 8\penalty0 (7):\penalty0 1182--1194, 2013.

\bibitem[DeepFloyd(2024)]{DeepFloyd2024}
DeepFloyd.
\newblock Deepfloyd.
\newblock \url{https://huggingface.co/DeepFloyd/IF-I-L-v1.0}, 2024.

\bibitem[Deng et~al.(2009)Deng, Dong, Socher, Li, Li, and Fei-Fei]{deng2009imagenet}
Jia Deng, Wei Dong, Richard Socher, Li-Jia Li, Kai Li, and Li Fei-Fei.
\newblock Imagenet: A large-scale hierarchical image database.
\newblock In \emph{2009 IEEE conference on computer vision and pattern recognition}, pages 248--255. Ieee, 2009.

\bibitem[Dhariwal and Nichol(2021)]{dhariwal2021guided}
Prafulla Dhariwal and Alexander Nichol.
\newblock Diffusion models beat gans on image synthesis.
\newblock \emph{Advances in neural information processing systems}, 34:\penalty0 8780--8794, 2021.

\bibitem[Dolhansky et~al.(2020)Dolhansky, Bitton, Pflaum, Lu, Howes, Wang, and Ferrer]{dolhansky2020deepfake}
Brian Dolhansky, Joanna Bitton, Ben Pflaum, Jikuo Lu, Russ Howes, Menglin Wang, and Cristian~Canton Ferrer.
\newblock The deepfake detection challenge (dfdc) dataset.
\newblock \emph{arXiv preprint arXiv:2006.07397}, 2020.

\bibitem[Dong et~al.(2013)Dong, Wang, and Tan]{dong2013casia}
Jing Dong, Wei Wang, and Tieniu Tan.
\newblock Casia image tampering detection evaluation database.
\newblock In \emph{2013 IEEE China summit and international conference on signal and information processing}, pages 422--426. IEEE, 2013.

\bibitem[Dosovitskiy et~al.(2021)Dosovitskiy, Beyer, Kolesnikov, Weissenborn, Zhai, Unterthiner, Dehghani, Minderer, Heigold, Gelly, Uszkoreit, and Houlsby]{dosovitskiy2020vit}
Alexey Dosovitskiy, Lucas Beyer, Alexander Kolesnikov, Dirk Weissenborn, Xiaohua Zhai, Thomas Unterthiner, Mostafa Dehghani, Matthias Minderer, Georg Heigold, Sylvain Gelly, Jakob Uszkoreit, and Neil Houlsby.
\newblock An image is worth 16x16 words: Transformers for image recognition at scale.
\newblock \emph{ICLR}, 2021.

\bibitem[Du et~al.(2021)Du, Li, Tenenbaum, and Mordatch]{du2021contrastiveEBM}
Yilun Du, Shuang Li, Joshua Tenenbaum, and Igor Mordatch.
\newblock Improved contrastive divergence training of energy-based models.
\newblock In \emph{International Conference on Machine Learning}, pages 2837--2848. PMLR, 2021.

\bibitem[Durall et~al.(2020)Durall, Keuper, and Keuper]{durall2020cnnSpectral}
Ricard Durall, Margret Keuper, and Janis Keuper.
\newblock Watch your up-convolution: Cnn based generative deep neural networks are failing to reproduce spectral distributions.
\newblock In \emph{Proceedings of the IEEE/CVF conference on computer vision and pattern recognition}, pages 7890--7899, 2020.

\bibitem[Dzanic et~al.(2020)Dzanic, Shah, and Witherden]{dzanic2020CNNfourier}
Tarik Dzanic, Karan Shah, and Freddie Witherden.
\newblock Fourier spectrum discrepancies in deep network generated images.
\newblock \emph{Advances in neural information processing systems}, 33:\penalty0 3022--3032, 2020.

\bibitem[Elflein et~al.(2021)Elflein, Charpentier, Zügner, and Günnemann]{elflein2021ebmOOD}
Sven Elflein, Bertrand Charpentier, Daniel Zügner, and Stephan Günnemann.
\newblock On out-of-distribution detection with energy-based models, 2021.

\bibitem[Epstein et~al.(2023)Epstein, Jain, Wang, and Zhang]{epstein2023online}
David~C Epstein, Ishan Jain, Oliver Wang, and Richard Zhang.
\newblock Online detection of ai-generated images.
\newblock In \emph{Proceedings of the IEEE/CVF International Conference on Computer Vision}, pages 382--392, 2023.

\bibitem[Esser et~al.(2021)Esser, Rombach, and Ommer]{esser2021taming}
Patrick Esser, Robin Rombach, and Bjorn Ommer.
\newblock Taming transformers for high-resolution image synthesis.
\newblock In \emph{Proceedings of the IEEE/CVF conference on computer vision and pattern recognition}, pages 12873--12883, 2021.

\bibitem[Face(2022)]{HuggingFaceDiffusersWeb}
Hugging Face.
\newblock Hugging face diffusers library.
\newblock \url{https://huggingface.co/models?library=diffusers}, accessed on June 05, 2022, 2022.

\bibitem[Gehrmann et~al.(2019)Gehrmann, Strobelt, and Rush]{gehrmann2019gltr}
Sebastian Gehrmann, Hendrik Strobelt, and Alexander~M Rush.
\newblock Gltr: Statistical detection and visualization of generated text.
\newblock \emph{arXiv preprint arXiv:1906.04043}, 2019.

\bibitem[Gloe and B{\"o}hme(2010)]{gloe2010dresden}
Thomas Gloe and Rainer B{\"o}hme.
\newblock The'dresden image database'for benchmarking digital image forensics.
\newblock In \emph{Proceedings of the 2010 ACM symposium on applied computing}, pages 1584--1590, 2010.

\bibitem[Goodfellow et~al.(2014)Goodfellow, Pouget-Abadie, Mirza, Xu, Warde-Farley, Ozair, Courville, and Bengio]{goodfellow2014generative}
Ian Goodfellow, Jean Pouget-Abadie, Mehdi Mirza, Bing Xu, David Warde-Farley, Sherjil Ozair, Aaron Courville, and Yoshua Bengio.
\newblock Generative adversarial nets.
\newblock \emph{Advances in Neural Information Processing Systems}, 27, 2014.

\bibitem[Google(2024)]{imagen3}
Google.
\newblock Imagen 3.
\newblock \url{https://deepmind.google/technologies/imagen-3}, 2024.

\bibitem[Gregor et~al.(2014)Gregor, Danihelka, Mnih, Blundell, and Wierstra]{gregor2014deep}
Karol Gregor, Ivo Danihelka, Andriy Mnih, Charles Blundell, and Daan Wierstra.
\newblock Deep autoregressive networks.
\newblock In \emph{International Conference on Machine Learning}, pages 1242--1250. PMLR, 2014.

\bibitem[Grommelt et~al.(2024)Grommelt, Weiss, Pfreundt, and Keuper]{grommelt2024fakeorjpeg}
Patrick Grommelt, Louis Weiss, Franz-Josef Pfreundt, and Janis Keuper.
\newblock Fake or jpeg? revealing common biases in generated image detection datasets.
\newblock \emph{arXiv preprint arXiv:2403.17608}, 2024.

\bibitem[Gu et~al.(2022)Gu, Chen, Bao, Wen, Zhang, Chen, Yuan, and Guo]{gu2022vqdiff}
Shuyang Gu, Dong Chen, Jianmin Bao, Fang Wen, Bo Zhang, Dongdong Chen, Lu Yuan, and Baining Guo.
\newblock Vector quantized diffusion model for text-to-image synthesis.
\newblock In \emph{Proceedings of the IEEE/CVF Conference on Computer Vision and Pattern Recognition}, pages 10696--10706, 2022.

\bibitem[Guillaro et~al.(2024)Guillaro, Zingarini, Usman, Sud, Cozzolino, and Verdoliva]{guillaro2024biasfreetrainingparadigmgeneral}
Fabrizio Guillaro, Giada Zingarini, Ben Usman, Avneesh Sud, Davide Cozzolino, and Luisa Verdoliva.
\newblock A bias-free training paradigm for more general ai-generated image detection, 2024.

\bibitem[Guo et~al.(2024)Guo, Asnani, Liu, and Liu]{guo2024tracing}
Xiao Guo, Vishal Asnani, Sijia Liu, and Xiaoming Liu.
\newblock Tracing hyperparameter dependencies for model parsing via learnable graph pooling network.
\newblock \emph{Advances in Neural Information Processing Systems}, 37:\penalty0 116899--116932, 2024.

\bibitem[Hadwiger and Riess(2021)]{hadwiger2021forchheim}
Benjamin Hadwiger and Christian Riess.
\newblock The forchheim image database for camera identification in the wild.
\newblock In \emph{Pattern Recognition. ICPR International Workshops and Challenges: Virtual Event, January 10--15, 2021, Proceedings, Part VI}, pages 500--515. Springer, 2021.

\bibitem[He et~al.(2016)He, Zhang, Ren, and Sun]{he2016resnet}
Kaiming He, Xiangyu Zhang, Shaoqing Ren, and Jian Sun.
\newblock Deep residual learning for image recognition.
\newblock In \emph{Proceedings of the IEEE conference on computer vision and pattern recognition}, pages 770--778, 2016.

\bibitem[Hendrycks and Gimpel(2016)]{hendrycks2016baseline}
Dan Hendrycks and Kevin Gimpel.
\newblock A baseline for detecting misclassified and out-of-distribution examples in neural networks.
\newblock In \emph{International Conference on Learning Representations}, 2016.

\bibitem[Hendrycks et~al.(2019)Hendrycks, Mazeika, Kadavath, and Song]{hendricks2019sslOOD}
Dan Hendrycks, Mantas Mazeika, Saurav Kadavath, and Dawn Song.
\newblock Using self-supervised learning can improve model robustness and uncertainty.
\newblock In \emph{Advances in Neural Information Processing Systems}. Curran Associates, Inc., 2019.

\bibitem[Hendrycks et~al.(2021)Hendrycks, Basart, Mu, Kadavath, Wang, Dorundo, Desai, Zhu, Parajuli, Guo, Song, Steinhardt, and Gilmer]{hendrycks2021robustOODdata}
Dan Hendrycks, Steven Basart, Norman Mu, Saurav Kadavath, Frank Wang, Evan Dorundo, Rahul Desai, Tyler Zhu, Samyak Parajuli, Mike Guo, Dawn Song, Jacob Steinhardt, and Justin Gilmer.
\newblock The many faces of robustness: A critical analysis of out-of-distribution generalization.
\newblock In \emph{Proceedings of the IEEE/CVF International Conference on Computer Vision (ICCV)}, pages 8340--8349, 2021.

\bibitem[Hong and Zhang(2024)]{hong2024wildfake}
Yan Hong and Jianfu Zhang.
\newblock Wildfake: A large-scale challenging dataset for ai-generated images detection.
\newblock \emph{arXiv preprint arXiv:2402.11843}, 2024.

\bibitem[Hou et~al.(2024)Hou, Ju, Sun, Jia, Ke, Zhou, Nikolich, and Lyu]{hou2024deepfake}
Shuwei Hou, Yan Ju, Chengzhe Sun, Shan Jia, Lipeng Ke, Riky Zhou, Anita Nikolich, and Siwei Lyu.
\newblock Deepfake-o-meter v2. 0: An open platform for deepfake detection.
\newblock \emph{arXiv preprint arXiv:2404.13146}, 2024.

\bibitem[Hu et~al.(2021)Hu, Shen, Wallis, Allen-Zhu, Li, Wang, Wang, and Chen]{hu2021lora}
Edward~J Hu, Yelong Shen, Phillip Wallis, Zeyuan Allen-Zhu, Yuanzhi Li, Shean Wang, Lu Wang, and Weizhu Chen.
\newblock Lora: Low-rank adaptation of large language models.
\newblock \emph{arXiv preprint arXiv:2106.09685}, 2021.

\bibitem[Hudson and Zitnick(2021)]{hudson2021gansformer}
Drew~A Hudson and Larry Zitnick.
\newblock Generative adversarial transformers.
\newblock In \emph{International conference on machine learning}, pages 4487--4499. PMLR, 2021.

\bibitem[Huh et~al.(2018)Huh, Liu, Owens, and Efros]{huh2018fighting}
Minyoung Huh, Andrew Liu, Andrew Owens, and Alexei~A Efros.
\newblock Fighting fake news: Image splice detection via learned self-consistency.
\newblock In \emph{Proceedings of the European conference on computer vision (ECCV)}, 2018.

\bibitem[Ilharco et~al.(2021)Ilharco, Wortsman, Wightman, Gordon, Carlini, Taori, Dave, Shankar, Namkoong, Miller, Hajishirzi, Farhadi, and Schmidt]{Ilharco2021OpenCLIP}
Gabriel Ilharco, Mitchell Wortsman, Ross Wightman, Cade Gordon, Nicholas Carlini, Rohan Taori, Achal Dave, Vaishaal Shankar, Hongseok Namkoong, John Miller, Hannaneh Hajishirzi, Ali Farhadi, and Ludwig Schmidt.
\newblock Openclip, 2021.
\newblock If you use this software, please cite it as below.

\bibitem[Jawahar et~al.(2020)Jawahar, Abdul-Mageed, and Lakshmanan]{jawahar2020automatic}
Ganesh Jawahar, Muhammad Abdul-Mageed, and Laks~VS Lakshmanan.
\newblock Automatic detection of machine generated text: A critical survey.
\newblock \emph{arXiv preprint arXiv:2011.01314}, 2020.

\bibitem[Johnson and Farid(2007)]{johnson2007exposing}
Micah~K Johnson and Hany Farid.
\newblock Exposing digital forgeries in complex lighting environments.
\newblock \emph{IEEE Transactions on Information Forensics and Security}, 2\penalty0 (3):\penalty0 450--461, 2007.

\bibitem[Kang et~al.(2023)Kang, Zhu, Zhang, Park, Shechtman, Paris, and Park]{kang2023gigagan}
Minguk Kang, Jun-Yan Zhu, Richard Zhang, Jaesik Park, Eli Shechtman, Sylvain Paris, and Taesung Park.
\newblock Scaling up gans for text-to-image synthesis.
\newblock In \emph{Proceedings of the IEEE Conference on Computer Vision and Pattern Recognition (CVPR)}, 2023.

\bibitem[Karras et~al.(2018)Karras, Aila, Laine, and Lehtinen]{karras2018proGAN}
Tero Karras, Timo Aila, Samuli Laine, and Jaakko Lehtinen.
\newblock Progressive growing of gans for improved quality, stability, and variation.
\newblock In \emph{International Conference on Learning Representations}, 2018.

\bibitem[Karras et~al.(2019)Karras, Laine, and Aila]{karras2019styleGAN}
Tero Karras, Samuli Laine, and Timo Aila.
\newblock A style-based generator architecture for generative adversarial networks.
\newblock In \emph{Proceedings of the IEEE/CVF conference on computer vision and pattern recognition}, pages 4401--4410, 2019.

\bibitem[Karras et~al.(2020{\natexlab{a}})Karras, Aittala, Hellsten, Laine, Lehtinen, and Aila]{karras2020stylegan2ada}
Tero Karras, Miika Aittala, Janne Hellsten, Samuli Laine, Jaakko Lehtinen, and Timo Aila.
\newblock Training generative adversarial networks with limited data.
\newblock \emph{Advances in neural information processing systems}, 33:\penalty0 12104--12114, 2020{\natexlab{a}}.

\bibitem[Karras et~al.(2020{\natexlab{b}})Karras, Laine, Aittala, Hellsten, Lehtinen, and Aila]{karras2020stylegan2}
Tero Karras, Samuli Laine, Miika Aittala, Janne Hellsten, Jaakko Lehtinen, and Timo Aila.
\newblock Analyzing and improving the image quality of stylegan.
\newblock In \emph{Proceedings of the IEEE/CVF conference on computer vision and pattern recognition}, pages 8110--8119, 2020{\natexlab{b}}.

\bibitem[Karras et~al.(2021)Karras, Aittala, Laine, H{\"a}rk{\"o}nen, Hellsten, Lehtinen, and Aila]{karras2021stylegan3}
Tero Karras, Miika Aittala, Samuli Laine, Erik H{\"a}rk{\"o}nen, Janne Hellsten, Jaakko Lehtinen, and Timo Aila.
\newblock Alias-free generative adversarial networks.
\newblock \emph{Advances in neural information processing systems}, 34:\penalty0 852--863, 2021.

\bibitem[Khalid et~al.(2021)Khalid, Tariq, Kim, and Woo]{khalid2021fakeavceleb}
Hasam Khalid, Shahroz Tariq, Minha Kim, and Simon Woo.
\newblock Fakeavceleb: A novel audio-video multimodal deepfake dataset.
\newblock In \emph{Proceedings of the Neural Information Processing Systems Track on Datasets and Benchmarks}, 2021.

\bibitem[Korus and Huang(2017)]{Korus2016TIFS}
P. Korus and J. Huang.
\newblock Multi-scale analysis strategies in prnu-based tampering localization.
\newblock \emph{IEEE Trans. on Information Forensics \& Security}, 2017.

\bibitem[Krishna et~al.(2024)Krishna, Song, Karpinska, Wieting, and Iyyer]{krishna2024paraphrasing}
Kalpesh Krishna, Yixiao Song, Marzena Karpinska, John Wieting, and Mohit Iyyer.
\newblock Paraphrasing evades detectors of ai-generated text, but retrieval is an effective defense.
\newblock \emph{Advances in Neural Information Processing Systems}, 36, 2024.

\bibitem[Kvikontent(2023)]{KvikontentMidjourney}
Kvikontent.
\newblock Kvikontent-midjourney v6.
\newblock https://huggingface.co/Kvikontent/midjourney-v6, 2023.

\bibitem[Kwon et~al.(2021)Kwon, You, Nam, Park, and Chae]{kwon2021kodf}
Patrick Kwon, Jaeseong You, Gyuhyeon Nam, Sungwoo Park, and Gyeongsu Chae.
\newblock Kodf: A large-scale korean deepfake detection dataset.
\newblock In \emph{Proceedings of the IEEE/CVF International Conference on Computer Vision}, 2021.

\bibitem[Labs(2024)]{flux}
Black~Forst Labs.
\newblock Flux.
\newblock \url{https://blackforestlabs.ai}, 2024.

\bibitem[Li et~al.(2022{\natexlab{a}})Li, Li, Xiong, and Hoi]{li2022blip}
Junnan Li, Dongxu Li, Caiming Xiong, and Steven Hoi.
\newblock Blip: Bootstrapping language-image pre-training for unified vision-language understanding and generation.
\newblock In \emph{ICML}, 2022{\natexlab{a}}.

\bibitem[Li et~al.(2019)Li, Zhang, and Malik]{li2019imle}
Ke Li, Tianhao Zhang, and Jitendra Malik.
\newblock Diverse image synthesis from semantic layouts via conditional imle.
\newblock In \emph{Proceedings of the IEEE/CVF International Conference on Computer Vision}, pages 4220--4229, 2019.

\bibitem[Li et~al.(2020)Li, Yang, Sun, Qi, and Lyu]{li2020celeb}
Yuezun Li, Xin Yang, Pu Sun, Honggang Qi, and Siwei Lyu.
\newblock Celeb-df: A large-scale challenging dataset for deepfake forensics.
\newblock In \emph{Proceedings of the IEEE/CVF conference on computer vision and pattern recognition}, 2020.

\bibitem[Li et~al.(2022{\natexlab{b}})Li, Wang, Xia, Liu, An, et~al.]{li2022likelihoodOOD}
Yewen Li, Chaojie Wang, Xiaobo Xia, Tongliang Liu, Bo An, et~al.
\newblock Out-of-distribution detection with an adaptive likelihood ratio on informative hierarchical vae.
\newblock \emph{Advances in Neural Information Processing Systems}, 35:\penalty0 7383--7396, 2022{\natexlab{b}}.

\bibitem[Liang et~al.(2018)Liang, Li, and Srikant]{liang2018temperatureOOD}
Shiyu Liang, Yixuan Li, and R Srikant.
\newblock Enhancing the reliability of out-of-distribution image detection in neural networks.
\newblock In \emph{International Conference on Learning Representations}, 2018.

\bibitem[Lin et~al.(2014)Lin, Maire, Belongie, Hays, Perona, Ramanan, Doll{\'a}r, and Zitnick]{lin2014mscoco}
Tsung-Yi Lin, Michael Maire, Serge Belongie, James Hays, Pietro Perona, Deva Ramanan, Piotr Doll{\'a}r, and C~Lawrence Zitnick.
\newblock Microsoft coco: Common objects in context.
\newblock In \emph{Computer Vision--ECCV 2014: 13th European Conference, Zurich, Switzerland, September 6-12, 2014, Proceedings, Part V 13}, pages 740--755. Springer, 2014.

\bibitem[Liu et~al.(2020)Liu, Wang, Owens, and Li]{liu2020energyOOD}
Weitang Liu, Xiaoyun Wang, John Owens, and Yixuan Li.
\newblock Energy-based out-of-distribution detection.
\newblock In \emph{Advances in Neural Information Processing Systems}, pages 21464--21475. Curran Associates, Inc., 2020.

\bibitem[Liu et~al.(2015)Liu, Luo, Wang, and Tang]{liu2015celeba}
Ziwei Liu, Ping Luo, Xiaogang Wang, and Xiaoou Tang.
\newblock Deep learning face attributes in the wild.
\newblock In \emph{Proceedings of International Conference on Computer Vision (ICCV)}, 2015.

\bibitem[Liu et~al.(2022)Liu, Mao, Wu, Feichtenhofer, Darrell, and Xie]{liu2022convnext}
Zhuang Liu, Hanzi Mao, Chao-Yuan Wu, Christoph Feichtenhofer, Trevor Darrell, and Saining Xie.
\newblock A convnet for the 2020s.
\newblock In \emph{Proceedings of the IEEE/CVF conference on computer vision and pattern recognition}, pages 11976--11986, 2022.

\bibitem[Loshchilov and Hutter(2019)]{loshchilov2018decoupled}
Ilya Loshchilov and Frank Hutter.
\newblock Decoupled weight decay regularization.
\newblock In \emph{International Conference on Learning Representations}, 2019.

\bibitem[Luo et~al.(2023{\natexlab{a}})Luo, Tan, Huang, Li, and Zhao]{luo2023latent}
Simian Luo, Yiqin Tan, Longbo Huang, Jian Li, and Hang Zhao.
\newblock Latent consistency models: Synthesizing high-resolution images with few-step inference.
\newblock \emph{arXiv preprint arXiv:2310.04378}, 2023{\natexlab{a}}.

\bibitem[Luo et~al.(2023{\natexlab{b}})Luo, Tan, Patil, Gu, von Platen, Passos, Huang, Li, and Zhao]{luo2023lcmLora}
Simian Luo, Yiqin Tan, Suraj Patil, Daniel Gu, Patrick von Platen, Apolin{\'a}rio Passos, Longbo Huang, Jian Li, and Hang Zhao.
\newblock Lcm-lora: A universal stable-diffusion acceleration module.
\newblock \emph{arXiv preprint arXiv:2311.05556}, 2023{\natexlab{b}}.

\bibitem[Marra et~al.(2019)Marra, Gragnaniello, Verdoliva, and Poggi]{marra2019gansFingerprint}
Francesco Marra, Diego Gragnaniello, Luisa Verdoliva, and Giovanni Poggi.
\newblock Do gans leave artificial fingerprints?
\newblock In \emph{2019 IEEE conference on multimedia information processing and retrieval (MIPR)}, pages 506--511. IEEE, 2019.

\bibitem[McInnes et~al.(2018)McInnes, Healy, Saul, and Gro{\ss}berger]{mcinnes2018umap}
Leland McInnes, John Healy, Nathaniel Saul, and Lukas Gro{\ss}berger.
\newblock Umap: Uniform manifold approximation and projection.
\newblock \emph{Journal of Open Source Software}, 3\penalty0 (29), 2018.

\bibitem[Micikevicius et~al.(2018)Micikevicius, Narang, Alben, Diamos, Elsen, Garcia, Ginsburg, Houston, Kuchaiev, Venkatesh, et~al.]{micikevicius2018mixed}
Paulius Micikevicius, Sharan Narang, Jonah Alben, Gregory Diamos, Erich Elsen, David Garcia, Boris Ginsburg, Michael Houston, Oleksii Kuchaiev, Ganesh Venkatesh, et~al.
\newblock Mixed precision training.
\newblock In \emph{International Conference on Learning Representations}, 2018.

\bibitem[Midjourney(2022)]{midjourney2022}
Inc. Midjourney.
\newblock Midjourney.
\newblock \url{https://www.midjourney.com/home}, 2022.

\bibitem[Mitchell et~al.(2023)Mitchell, Lee, Khazatsky, Manning, and Finn]{mitchell2023detectgpt}
Eric Mitchell, Yoonho Lee, Alexander Khazatsky, Christopher~D Manning, and Chelsea Finn.
\newblock Detectgpt: Zero-shot machine-generated text detection using probability curvature.
\newblock In \emph{International Conference on Machine Learning}, pages 24950--24962. PMLR, 2023.

\bibitem[Mohseni et~al.(2020)Mohseni, Pitale, Yadawa, and Wang]{mohseni2020sslOOD}
Sina Mohseni, Mandar Pitale, JBS Yadawa, and Zhangyang Wang.
\newblock Self-supervised learning for generalizable out-of-distribution detection.
\newblock \emph{Proceedings of the AAAI Conference on Artificial Intelligence}, 34\penalty0 (04):\penalty0 5216--5223, 2020.

\bibitem[Ng et~al.(2004)Ng, Chang, and Sun]{ng2004data}
Tian-Tsong Ng, Shih-Fu Chang, and Q Sun.
\newblock A data set of authentic and spliced image blocks.
\newblock \emph{Columbia University, ADVENT Technical Report}, 4, 2004.

\bibitem[Nichol et~al.(2022)Nichol, Dhariwal, Ramesh, Shyam, Mishkin, Mcgrew, Sutskever, and Chen]{nichol2022glide}
Alexander~Quinn Nichol, Prafulla Dhariwal, Aditya Ramesh, Pranav Shyam, Pamela Mishkin, Bob Mcgrew, Ilya Sutskever, and Mark Chen.
\newblock Glide: Towards photorealistic image generation and editing with text-guided diffusion models.
\newblock In \emph{International Conference on Machine Learning}, pages 16784--16804. PMLR, 2022.

\bibitem[Novozamsky et~al.(2020)Novozamsky, Mahdian, and Saic]{novozamsky2020imd2020}
Adam Novozamsky, Babak Mahdian, and Stanislav Saic.
\newblock Imd2020: A large-scale annotated dataset tailored for detecting manipulated images.
\newblock In \emph{Proceedings of the IEEE/CVF Winter Conference on Applications of Computer Vision Workshops}, pages 71--80, 2020.

\bibitem[Ojha et~al.(2023)Ojha, Li, and Lee]{Ojha_2023_CVPR}
Utkarsh Ojha, Yuheng Li, and Yong~Jae Lee.
\newblock Towards universal fake image detectors that generalize across generative models.
\newblock In \emph{Proceedings of the IEEE/CVF Conference on Computer Vision and Pattern Recognition (CVPR)}, pages 24480--24489, 2023.

\bibitem[OpenAI(2022)]{dalle2}
OpenAI.
\newblock Dall-e 2.
\newblock \url{https://openai.com/index/dall-e-2}, 2022.

\bibitem[OpenAI(2023)]{dalle3}
OpenAI.
\newblock Dall-e 3.
\newblock \url{https://openai.com/index/dall-e-3}, 2023.

\bibitem[OpenRail-M~License()]{creativeml2022}
CreativeML OpenRail-M~License.
\newblock Creativeml openrail-m license.
\newblock \url{https://huggingface.co/spaces/CompVis/stable-diffusion-license}.

\bibitem[Park et~al.(2019)Park, Liu, Wang, and Zhu]{park2019gauGAN}
Taesung Park, Ming-Yu Liu, Ting-Chun Wang, and Jun-Yan Zhu.
\newblock Semantic image synthesis with spatially-adaptive normalization.
\newblock In \emph{Proceedings of the IEEE Conference on Computer Vision and Pattern Recognition}, 2019.

\bibitem[Paszke et~al.(2019)Paszke, Gross, Massa, Lerer, Bradbury, Chanan, Killeen, Lin, Gimelshein, Antiga, et~al.]{paszke2019pytorch}
Adam Paszke, Sam Gross, Francisco Massa, Adam Lerer, James Bradbury, Gregory Chanan, Trevor Killeen, Zeming Lin, Natalia Gimelshein, Luca Antiga, et~al.
\newblock Pytorch: An imperative style, high-performance deep learning library.
\newblock \emph{Advances in neural information processing systems}, 32, 2019.

\bibitem[Peebles and Xie(2023)]{peebles2023dit}
William Peebles and Saining Xie.
\newblock Scalable diffusion models with transformers.
\newblock In \emph{Proceedings of the IEEE/CVF International Conference on Computer Vision}, pages 4195--4205, 2023.

\bibitem[Pernias et~al.(2023)Pernias, Rampas, Richter, Pal, and Aubreville]{pernias2023stablecascade}
Pablo Pernias, Dominic Rampas, Mats~Leon Richter, Christopher Pal, and Marc Aubreville.
\newblock W{\"u}rstchen: An efficient architecture for large-scale text-to-image diffusion models.
\newblock In \emph{The Twelfth International Conference on Learning Representations}, 2023.

\bibitem[Podell et~al.(2023)Podell, English, Lacey, Blattmann, Dockhorn, M{\"u}ller, Penna, and Rombach]{podell2023sdxl}
Dustin Podell, Zion English, Kyle Lacey, Andreas Blattmann, Tim Dockhorn, Jonas M{\"u}ller, Joe Penna, and Robin Rombach.
\newblock Sdxl: Improving latent diffusion models for high-resolution image synthesis.
\newblock In \emph{The Twelfth International Conference on Learning Representations}, 2023.

\bibitem[Popescu and Farid(2005)]{popescu2005exposing}
Alin~C Popescu and Hany Farid.
\newblock Exposing digital forgeries by detecting traces of resampling.
\newblock \emph{IEEE Transactions on signal processing}, 2005.

\bibitem[Radford et~al.(2021)Radford, Kim, Hallacy, Ramesh, Goh, Agarwal, Sastry, Askell, Mishkin, Clark, et~al.]{radford2021clip}
Alec Radford, Jong~Wook Kim, Chris Hallacy, Aditya Ramesh, Gabriel Goh, Sandhini Agarwal, Girish Sastry, Amanda Askell, Pamela Mishkin, Jack Clark, et~al.
\newblock Learning transferable visual models from natural language supervision.
\newblock In \emph{International conference on machine learning}, pages 8748--8763. PMLR, 2021.

\bibitem[Rajan et~al.(2025)Rajan, Ojha, Schloesser, and Lee]{rajan2025aligneddatasetsimprovedetection}
Anirudh~Sundara Rajan, Utkarsh Ojha, Jedidiah Schloesser, and Yong~Jae Lee.
\newblock Aligned datasets improve detection of latent diffusion-generated images, 2025.

\bibitem[Ren et~al.(2019)Ren, Liu, Fertig, Snoek, Poplin, Depristo, Dillon, and Lakshminarayanan]{ren2019likelihood}
Jie Ren, Peter~J Liu, Emily Fertig, Jasper Snoek, Ryan Poplin, Mark Depristo, Joshua Dillon, and Balaji Lakshminarayanan.
\newblock Likelihood ratios for out-of-distribution detection.
\newblock \emph{Advances in neural information processing systems}, 32, 2019.

\bibitem[Ricker et~al.(2024)Ricker, Lukovnikov, and Fischer]{ricker2024aeroblade}
Jonas Ricker, Denis Lukovnikov, and Asja Fischer.
\newblock Aeroblade: Training-free detection of latent diffusion images using autoencoder reconstruction error.
\newblock In \emph{Proceedings of the IEEE/CVF Conference on Computer Vision and Pattern Recognition}, pages 9130--9140, 2024.

\bibitem[Rombach et~al.(2022)Rombach, Blattmann, Lorenz, Esser, and Ommer]{rombach2022LDM}
Robin Rombach, Andreas Blattmann, Dominik Lorenz, Patrick Esser, and Bj\"orn Ommer.
\newblock High-resolution image synthesis with latent diffusion models.
\newblock In \emph{Proceedings of the IEEE/CVF Conference on Computer Vision and Pattern Recognition (CVPR)}, pages 10684--10695, 2022.

\bibitem[R{\"o}ssler et~al.(2018)R{\"o}ssler, Cozzolino, Verdoliva, Riess, Thies, and Nie{\ss}ner]{rossler2018faceforensics}
Andreas R{\"o}ssler, Davide Cozzolino, Luisa Verdoliva, Christian Riess, Justus Thies, and Matthias Nie{\ss}ner.
\newblock Faceforensics: A large-scale video dataset for forgery detection in human faces.
\newblock \emph{arXiv preprint arXiv:1803.09179}, 2018.

\bibitem[Rossler et~al.(2019)Rossler, Cozzolino, Verdoliva, Riess, Thies, and Nie{\ss}ner]{rossler2019faceforensics++}
Andreas Rossler, Davide Cozzolino, Luisa Verdoliva, Christian Riess, Justus Thies, and Matthias Nie{\ss}ner.
\newblock Faceforensics++: Learning to detect manipulated facial images.
\newblock In \emph{Proceedings of the IEEE/CVF international conference on computer vision}, 2019.

\bibitem[Sadasivan et~al.(2023)Sadasivan, Kumar, Balasubramanian, Wang, and Feizi]{sadasivan2023can}
Vinu~Sankar Sadasivan, Aounon Kumar, Sriram Balasubramanian, Wenxiao Wang, and Soheil Feizi.
\newblock Can ai-generated text be reliably detected?
\newblock \emph{arXiv preprint arXiv:2303.11156}, 2023.

\bibitem[Sastry and Oore(2020)]{sastry2020oodGram}
Chandramouli~Shama Sastry and Sageev Oore.
\newblock Detecting out-of-distribution examples with {G}ram matrices.
\newblock In \emph{Proceedings of the 37th International Conference on Machine Learning}, pages 8491--8501. PMLR, 2020.

\bibitem[Sauer et~al.(2021)Sauer, Chitta, M{\"{u}}ller, and Geiger]{Sauer2021projectedGAN}
Axel Sauer, Kashyap Chitta, Jens M{\"{u}}ller, and Andreas Geiger.
\newblock Projected gans converge faster.
\newblock In \emph{Advances in Neural Information Processing Systems (NeurIPS)}, 2021.

\bibitem[Sauer et~al.(2022)Sauer, Schwarz, and Geiger]{sauer2022styleganxl}
Axel Sauer, Katja Schwarz, and Andreas Geiger.
\newblock Stylegan-xl: Scaling stylegan to large diverse datasets.
\newblock In \emph{ACM SIGGRAPH 2022 conference proceedings}, pages 1--10, 2022.

\bibitem[Schuhmann et~al.(2021)Schuhmann, Vencu, Beaumont, Kaczmarczyk, Mullis, Katta, Coombes, Jitsev, and Komatsuzaki]{schuhmann2021laion}
Christoph Schuhmann, Richard Vencu, Romain Beaumont, Robert Kaczmarczyk, Clayton Mullis, Aarush Katta, Theo Coombes, Jenia Jitsev, and Aran Komatsuzaki.
\newblock Laion-400m: Open dataset of clip-filtered 400 million image-text pairs.
\newblock \emph{arXiv preprint arXiv:2111.02114}, 2021.

\bibitem[Schuhmann et~al.(2022)Schuhmann, Beaumont, Vencu, Gordon, Wightman, Cherti, Coombes, Katta, Mullis, Wortsman, Schramowski, Kundurthy, Crowson, Schmidt, Kaczmarczyk, and Jitsev]{schuhmann2022laion5b}
Christoph Schuhmann, Romain Beaumont, Richard Vencu, Cade~W Gordon, Ross Wightman, Mehdi Cherti, Theo Coombes, Aarush Katta, Clayton Mullis, Mitchell Wortsman, Patrick Schramowski, Srivatsa~R Kundurthy, Katherine Crowson, Ludwig Schmidt, Robert Kaczmarczyk, and Jenia Jitsev.
\newblock {LAION}-5b: An open large-scale dataset for training next generation image-text models.
\newblock In \emph{Thirty-sixth Conference on Neural Information Processing Systems Datasets and Benchmarks Track}, 2022.

\bibitem[Sehwag et~al.(2020)Sehwag, Chiang, and Mittal]{sehwag2020ssd}
Vikash Sehwag, Mung Chiang, and Prateek Mittal.
\newblock Ssd: A unified framework for self-supervised outlier detection.
\newblock In \emph{International Conference on Learning Representations}, 2020.

\bibitem[Shakhmatov et~al.(2023)Shakhmatov, Razzhigaev, Nikolich, Arkhipkin, Pavlov, Kuznetsov, and Dimitrov]{kandinsky2023models}
Arseniy Shakhmatov, Anton Razzhigaev, Aleksandr Nikolich, Vladimir Arkhipkin, Igor Pavlov, Andrey Kuznetsov, and Denis Dimitrov.
\newblock Kandinsky 2.2.
\newblock \url{https://github.com/ai-forever/Kandinsky-2}, 2023.

\bibitem[Shullani et~al.(2017)Shullani, Fontani, Iuliani, Shaya, and Piva]{shullani2017vision}
Dasara Shullani, Marco Fontani, Massimo Iuliani, Omar~Al Shaya, and Alessandro Piva.
\newblock Vision: a video and image dataset for source identification.
\newblock \emph{EURASIP Journal on Information Security}, 2017:\penalty0 1--16, 2017.

\bibitem[Skorokhodov et~al.(2021)Skorokhodov, Sotnikov, and Elhoseiny]{Skorokhodov2021landscapes}
Ivan Skorokhodov, Grigorii Sotnikov, and Mohamed Elhoseiny.
\newblock Aligning latent and image spaces to connect the unconnectable.
\newblock \emph{arXiv preprint arXiv:2104.06954}, 2021.

\bibitem[Solaiman et~al.(2019)Solaiman, Brundage, Clark, Askell, Herbert-Voss, Wu, Radford, Krueger, Kim, Kreps, et~al.]{solaiman2019release}
Irene Solaiman, Miles Brundage, Jack Clark, Amanda Askell, Ariel Herbert-Voss, Jeff Wu, Alec Radford, Gretchen Krueger, Jong~Wook Kim, Sarah Kreps, et~al.
\newblock Release strategies and the social impacts of language models.
\newblock \emph{arXiv preprint arXiv:1908.09203}, 2019.

\bibitem[Song et~al.(2023)Song, Dhariwal, Chen, and Sutskever]{song2023consistency}
Yang Song, Prafulla Dhariwal, Mark Chen, and Ilya Sutskever.
\newblock Consistency models.
\newblock In \emph{Proceedings of the 40th International Conference on Machine Learning}, pages 32211--32252, 2023.

\bibitem[Takida et~al.(2024)Takida, Imaizumi, Shibuya, Lai, Uesaka, Murata, and Mitsufuji]{takida2024san}
Yuhta Takida, Masaaki Imaizumi, Takashi Shibuya, Chieh-Hsin Lai, Toshimitsu Uesaka, Naoki Murata, and Yuki Mitsufuji.
\newblock {SAN}: Inducing metrizability of {GAN} with discriminative normalized linear layer.
\newblock In \emph{The Twelfth International Conference on Learning Representations}, 2024.

\bibitem[Tan et~al.(2023)Tan, Zhao, Wei, Gu, and Wei]{tan2023learning}
Chuangchuang Tan, Yao Zhao, Shikui Wei, Guanghua Gu, and Yunchao Wei.
\newblock Learning on gradients: Generalized artifacts representation for gan-generated images detection.
\newblock In \emph{Proceedings of the IEEE/CVF Conference on Computer Vision and Pattern Recognition}, pages 12105--12114, 2023.

\bibitem[Tan et~al.(2024)Tan, Zhao, Wei, Gu, Liu, and Wei]{tan2024rethinking}
Chuangchuang Tan, Yao Zhao, Shikui Wei, Guanghua Gu, Ping Liu, and Yunchao Wei.
\newblock Rethinking the up-sampling operations in cnn-based generative network for generalizable deepfake detection.
\newblock In \emph{Proceedings of the IEEE/CVF Conference on Computer Vision and Pattern Recognition}, pages 28130--28139, 2024.

\bibitem[Tao et~al.(2022)Tao, Tang, Wu, Jing, Bao, and Xu]{tao2022df}
Ming Tao, Hao Tang, Fei Wu, Xiao-Yuan Jing, Bing-Kun Bao, and Changsheng Xu.
\newblock Df-gan: A simple and effective baseline for text-to-image synthesis.
\newblock In \emph{Proceedings of the IEEE/CVF conference on computer vision and pattern recognition}, pages 16515--16525, 2022.

\bibitem[Tao et~al.(2023)Tao, Bao, Tang, and Xu]{tao2023galip}
Ming Tao, Bing-Kun Bao, Hao Tang, and Changsheng Xu.
\newblock Galip: Generative adversarial clips for text-to-image synthesis.
\newblock In \emph{Proceedings of the IEEE/CVF Conference on Computer Vision and Pattern Recognition}, pages 14214--14223, 2023.

\bibitem[Team(2024)]{DeciFoundationModels}
DeciAI~Research Team.
\newblock Decidiffusion 2.0, 2024.

\bibitem[Uchendu et~al.(2020)Uchendu, Le, Shu, and Lee]{uchendu2020authorship}
Adaku Uchendu, Thai Le, Kai Shu, and Dongwon Lee.
\newblock Authorship attribution for neural text generation.
\newblock In \emph{Proceedings of the 2020 Conference on Empirical Methods in Natural Language Processing (EMNLP)}, pages 8384--8395, 2020.

\bibitem[Van Den~Oord et~al.(2017)Van Den~Oord, Vinyals, et~al.]{van2017vqvae}
Aaron Van Den~Oord, Oriol Vinyals, et~al.
\newblock Neural discrete representation learning.
\newblock \emph{Advances in neural information processing systems}, 30, 2017.

\bibitem[von Platen et~al.(2022)von Platen, Patil, Lozhkov, Cuenca, Lambert, Rasul, Davaadorj, Nair, Paul, Berman, Xu, Liu, and Wolf]{von-platen-etal-2022-diffusers}
Patrick von Platen, Suraj Patil, Anton Lozhkov, Pedro Cuenca, Nathan Lambert, Kashif Rasul, Mishig Davaadorj, Dhruv Nair, Sayak Paul, William Berman, Yiyi Xu, Steven Liu, and Thomas Wolf.
\newblock Diffusers: State-of-the-art diffusion models.
\newblock \url{https://github.com/huggingface/diffusers}, 2022.

\bibitem[Vyas et~al.(2018)Vyas, Jammalamadaka, Zhu, Das, Kaul, and Willke]{vyas2018sslOOD}
Apoorv Vyas, Nataraj Jammalamadaka, Xia Zhu, Dipankar Das, Bharat Kaul, and Theodore~L. Willke.
\newblock Out-of-distribution detection using an ensemble of self supervised leave-out classifiers.
\newblock In \emph{Proceedings of the European Conference on Computer Vision (ECCV)}, 2018.

\bibitem[Wang et~al.(2020)Wang, Wang, Zhang, Owens, and Efros]{Wang_2020_CVPR}
Sheng-Yu Wang, Oliver Wang, Richard Zhang, Andrew Owens, and Alexei~A. Efros.
\newblock Cnn-generated images are surprisingly easy to spot... for now.
\newblock In \emph{Proceedings of the IEEE/CVF Conference on Computer Vision and Pattern Recognition (CVPR)}, 2020.

\bibitem[Wightman(2019)]{rw2019timm}
Ross Wightman.
\newblock Pytorch image models.
\newblock \url{https://github.com/huggingface/pytorch-image-models}, 2019.

\bibitem[Wukong(2022)]{wukong2022}
Wukong.
\newblock \url{https://xihe.mindspore.cn/modelzoo/wukong}, 2022.

\bibitem[Xiao et~al.(2020)Xiao, Yan, and Amit]{xiao2020likelihoodRegretOOD}
Zhisheng Xiao, Qing Yan, and Yali Amit.
\newblock Likelihood regret: An out-of-distribution detection score for variational auto-encoder.
\newblock In \emph{Advances in Neural Information Processing Systems}, pages 20685--20696. Curran Associates, Inc., 2020.

\bibitem[Yu et~al.(2015)Yu, Zhang, Song, Seff, and Xiao]{yu15lsun}
Fisher Yu, Yinda Zhang, Shuran Song, Ari Seff, and Jianxiong Xiao.
\newblock Lsun: Construction of a large-scale image dataset using deep learning with humans in the loop.
\newblock \emph{arXiv preprint arXiv:1506.03365}, 2015.

\bibitem[Zhang et~al.(2022)Zhang, Gu, Zhang, Bao, Chen, Wen, Wang, and Guo]{zhang2022styleswin}
Bowen Zhang, Shuyang Gu, Bo Zhang, Jianmin Bao, Dong Chen, Fang Wen, Yong Wang, and Baining Guo.
\newblock Styleswin: Transformer-based gan for high-resolution image generation.
\newblock In \emph{Proceedings of the IEEE/CVF conference on computer vision and pattern recognition}, pages 11304--11314, 2022.

\bibitem[Zhang et~al.(2019)Zhang, Karaman, and Chang]{zhang2019GANartifacts}
Xu Zhang, Svebor Karaman, and Shih-Fu Chang.
\newblock Detecting and simulating artifacts in gan fake images.
\newblock In \emph{2019 IEEE international workshop on information forensics and security (WIFS)}, pages 1--6. IEEE, 2019.

\bibitem[Zhu et~al.(2017)Zhu, Park, Isola, and Efros]{zhu2017cycleGAN}
Jun-Yan Zhu, Taesung Park, Phillip Isola, and Alexei~A Efros.
\newblock Unpaired image-to-image translation using cycle-consistent adversarial networks.
\newblock In \emph{Computer Vision (ICCV), 2017 IEEE International Conference on}, 2017.

\bibitem[Zhu et~al.(2023)Zhu, Chen, YAN, Huang, Lin, Li, Tu, Hu, Hu, and Wang]{zhu2024genimage}
Mingjian Zhu, Hanting Chen, Qiangyu YAN, Xudong Huang, Guanyu Lin, Wei Li, Zhijun Tu, Hailin Hu, Jie Hu, and Yunhe Wang.
\newblock Genimage: A million-scale benchmark for detecting ai-generated image.
\newblock In \emph{Advances in Neural Information Processing Systems Dataset and Benchmarks Track}, pages 77771--77782, 2023.

\bibitem[Zi et~al.(2020)Zi, Chang, Chen, Ma, and Jiang]{zi2020wilddeepfake}
Bojia Zi, Minghao Chang, Jingjing Chen, Xingjun Ma, and Yu-Gang Jiang.
\newblock Wilddeepfake: A challenging real-world dataset for deepfake detection.
\newblock In \emph{Proceedings of the 28th ACM international conference on multimedia}, 2020.

\end{thebibliography}
}
\supparxiv{}{\clearpage
\supparxiv{\setcounter{page}{1}}{}
\supparxiv{\setcounter{figure}{7}}{}
\supparxiv{\setcounter{table}{2}}{}
\supparxiv{\maketitlesupplementary}{}
\appendix

\section{Other applications}
\label{appendix:other_applications}
\vspace{-8pt}
\begin{figure}[ht]
    \centering
    \begin{subfigure}{0.48\linewidth}
        \includegraphics[width=\linewidth]{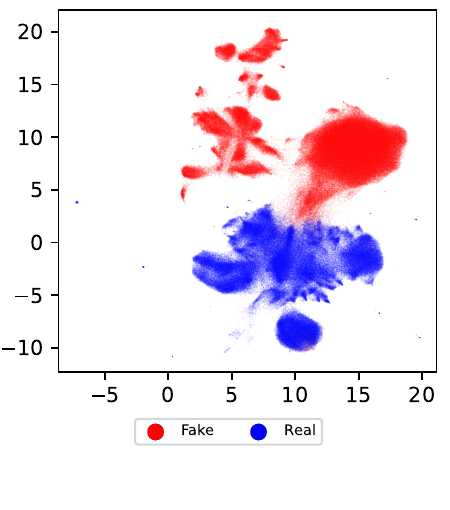}
        \caption{Fake \vs{} real visualization}
        \label{fig_apdx:fake_real_feature_visualization}
    \end{subfigure}
    \begin{subfigure}{0.48\linewidth}
        \includegraphics[width=\linewidth]{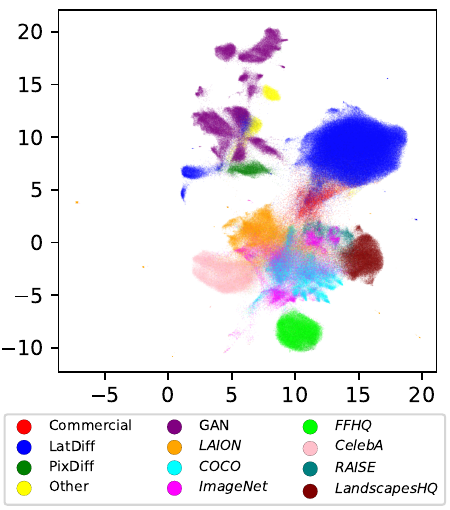}
        \caption{Generator type visualization}
        \label{fig_apdx:generator_type_feature_visualization}
    \end{subfigure}\vspace{-2pt}
    \caption{\textbf{Feature space visualization.} We visualize a feature space of our trained classifier using 10\% of our training data and the evaluation set. For better visibility, only a subset of our real datasets are visualized and the labels for real datasets are italicized. We observe a good separation between \textit{fake} \vs{} \textit{real} data, and between different generator types and real datasets.}\supparxiv{\vspace{-1mm}}{}
    \label{fig_apdx:feature_visualization}
\end{figure}

Other applications, beyond the ``real-or-fake'' image forensics task, could potentially be supported by our dataset.
In particular, a diverse array of generators and their corresponding images in our dataset may be valuable for addressing the \textit{generator attribution} problem, where the goal is to identify the characteristics of the underlying generator that is responsible for synthesizing a given image.

\Cref{fig_apdx:feature_visualization} presents a UMAP~\cite{mcinnes2018umap} visualization of the feature space of our trained classifier. 
We use the activation of the penultimate layer for visualization following Ojha \etal{}~\cite{Ojha_2023_CVPR}. 
The feature space reveals interesting structure: GANs form a clearly separated cluster; most commercial models are distributed closely to latent diffusion models; real datasets such as LAION~\cite{schuhmann2021laion}, ImageNet~\cite{deng2009imagenet}, COCO~\cite{lin2014mscoco}, and RAISE~\cite{dang2015raise} are closely distributed, whereas CelebA~\cite{liu2015celeba}, FFHQ~\cite{karras2019styleGAN}, and Landscapes HQ~\cite{Skorokhodov2021landscapes} appear to be more isolated.
It is important to note that these separations emerge naturally without explicit training. A targeted learning objective may further enhance these separations.

Building on the feature space observations, we use a $k$-nearest-neighbor classifier with $k\!=\!5$ using 10\% of our training data to identify the generator types in our evaluation set.
We separate generators as ``known'' (i.e., GANs, latent and pixel diffusions, and real data) and ``unknown'' (commercial models and Stable Cascade~\cite{pernias2023stablecascade}) generator types and compute the confusion matrices as shown in \Cref{fig_apdx:confusion_matrix}.
Note that none of these generators are seen during training.
\Cref{fig_apdx:confusion_marix_known} demonstrates strong performance in identifying GANs, latent diffusion models, and real data. However, pixel-based diffusion models show lower performance, possibly due to their limited representation (only 3 models) in our training set.
The classification result for the ``unknown'' set is shown in \Cref{fig_apdx:confusion_marix_unknown}.
Interestingly, commercial models are predominantly classified as latent diffusion or GANs, while Stable Cascade~\cite{pernias2023stablecascade} displays similarity to latent diffusion models despite their unique three-stage sampling process. %

\begin{figure}[t]
    \centering
    \begin{subfigure}[b]{0.48\linewidth}
        \includegraphics[width=\linewidth]{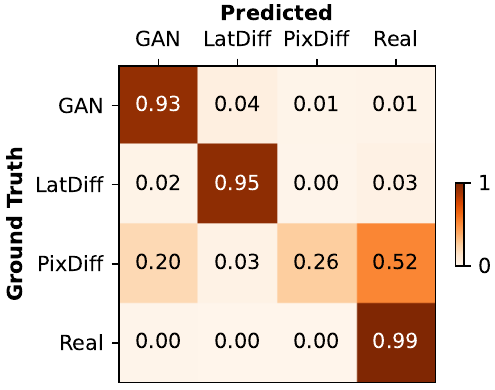}
        \caption{Known architectures}
        \label{fig_apdx:confusion_marix_known}
    \end{subfigure}%
    \hfill
    \begin{subfigure}[b]{0.48\linewidth}
        \includegraphics[width=\linewidth]{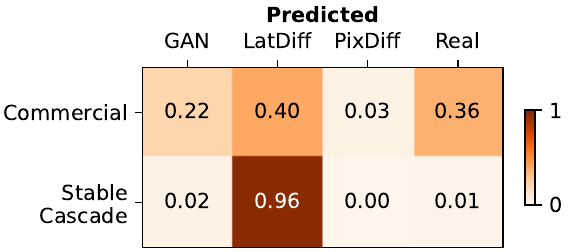}
        \vspace{6pt}
        \caption{Unknown architectures}
        \label{fig_apdx:confusion_marix_unknown}
    \end{subfigure}
    \caption{\textbf{Generator type classification.} We classify the generator type of a given image using $k$-nearest-neighbor. (a) Confusion matrix of ``known'' generator types. We observe high accuracy in GANs, latent diffusions, and real data. (b) Classification results on ``unknown'' architectures. Commercial models are predominantly classified as latent diffusion and GANs (disregarding `real'). Stable Cascade~\cite{pernias2023stablecascade}, which we categorized as \textit{Other} generator type, shows similarity to latent diffusion models.}\supparxiv{\figvspace{}}{}
    \label{fig_apdx:confusion_matrix}
\end{figure}

\section{Dataset composition}
\vspace{9pt}
\paragraph{Generator licenses.}
\label{appendix:generator_license_composition}

In \Cref{fig_apdx:modelLicense}, we report the generator licenses in our dataset. Most of the models use the \texttt{CreativeML OpenRAIL-M} license~\cite{creativeml2022}.

\paragraph{Model metadata.}
\label{appendix:model_metadata}
We show an example model metadata in \cref{tab_apdx:model_metadata}. It contains the name of the models, their categorized architectures, licenses, source real datasets, and the Hugging Face tags if available.

\paragraph{Model composition.} 
The composition of the training set of \datasetName{} is detailed in \Cref{tab_apdx:modelcounts,fig_apdx:imagecounts}.
A vast majority of the models and generated images are latent diffusion.
\Cref{fig_apdx:evalset_composition} illustrates the composition of the evaluation set, which includes two variants of HDiT~\cite{crowson2024scalable}: one trained on FFHQ~\cite{karras2019styleGAN} and another on ImageNet~\cite{deng2009imagenet}.
For computing metrics such as mAP and accuracy, these HDiT variants are treated as separate entities due to their distinct training data and model weights. However, when reporting the number of models in our dataset, we count them as a single model. 

\begin{table*}[th!]
\centering
\scriptsize
\ttfamily  %
\setlength{\tabcolsep}{4pt}  %
\adjustbox{max width=\supparxiv{0.95\linewidth}{\linewidth}}
{
    \begin{tabularx}{\linewidth}{|>{\raggedright\arraybackslash}p{1.8cm}|>{\raggedright\arraybackslash}p{2cm}|>{\raggedright\arraybackslash}p{1.8cm}|>{\raggedright\arraybackslash}X|>{\raggedright\arraybackslash}p{2.5cm}|>{\raggedright\arraybackslash}p{2.5cm}|}
    \hline
    \textbf{\makecell[c]{\textbf{Model}}} & \textbf{\makecell[c]{\textbf{Architecture}}} & \textbf{\makecell[c]{\textbf{License}}} & \textbf{\makecell[c]{\textbf{RealSource}}} & \textbf{\makecell[c]{\textbf{HF\_pipeline\_tag}}} & \textbf{\makecell[c]{\textbf{HF\_diffusers\_tag}}} \\
    \hline
    \makecell[l]{danbochman/\\ccxl} & LatentDiff & None & \makecell[l]{coco,forchheim,imagenet,imd2020,laion,\\landscapesHQ,vision} & \makecell[l]{StableDiffusionXL-\\Pipeline} & \makecell[l]{StableDiffusionXL-\\Pipeline} \\
    \hline
    \makecell[l]{livingbox/\\modern-\\style-v3} & LatentDiff & \makecell[l]{creativeml-\\openrail-m} & \makecell[l]{coco,forchheim,imagenet,imd2020,laion,\\landscapesHQ,vision} & \makecell[l]{StableDiffusion-\\Pipeline} & stable-diffusion \\
    \hline
    \multicolumn{6}{|c|}{\ldots} \\
    \hline
    DeepFloyd & PixelDiff & DeepFloyd-IF & coco & N/A & N/A \\
    \hline
    BigGAN & GAN & MIT & imagenet & N/A & N/A \\
    \hline
    \multicolumn{6}{|c|}{\ldots} \\
    \hline
    \end{tabularx}
}
\supparxiv{\vspace{-1mm}}{}
\caption{\textbf{Example model metadata.} We log both the author and model names for the Hugging Face~\cite{HuggingFaceDiffusersWeb} models and only the model names for others. We also log the generator type (i.e., architecture), model license, source real dataset, and Hugging Face tags if available.}\figvspace{}
\label{tab_apdx:model_metadata}
\end{table*}

\begin{figure}[tbh]
    \centering
    \includegraphics[width=\linewidth]{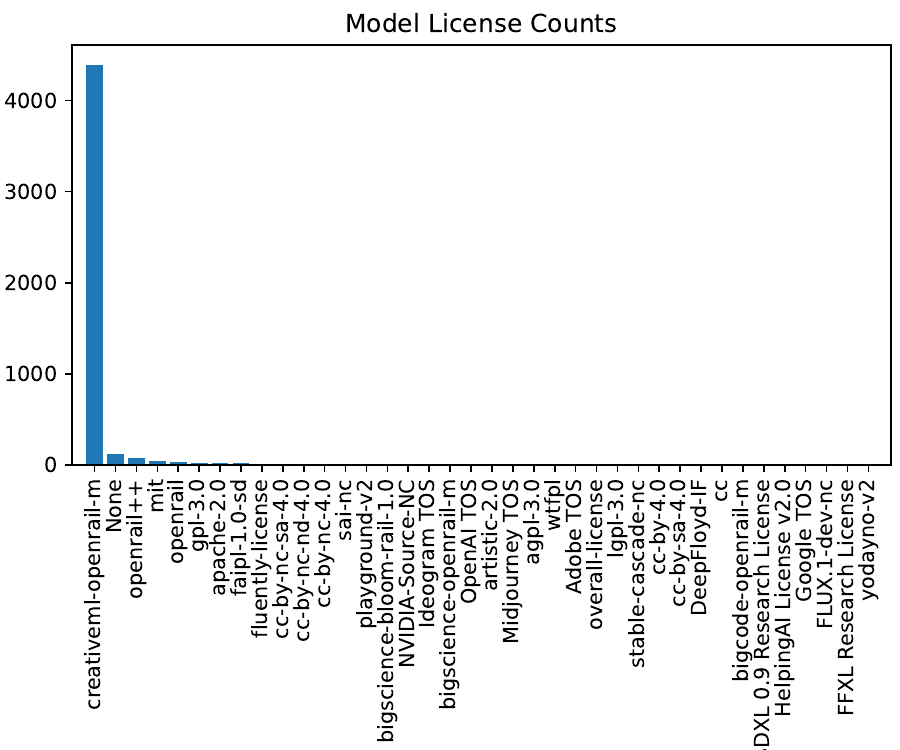}
    \supparxiv{\vspace{-1mm}}{}
    \caption{Histogram of model licenses in our dataset. A vast majority of the models use the \texttt{CreativeML OpenRAIL-M} license~\cite{creativeml2022}.}
    \label{fig_apdx:modelLicense}
\end{figure}

\begin{table}[t]
    \centering
    \adjustbox{max width=\supparxiv{0.95\linewidth}{\linewidth}}{
    \begin{tabular}{lrrrr}
        \toprule
         & Latent Diff. & GAN & Pixel Diff. & Other \\
        \midrule
        Models & 4766 & 12 & 3 & 1 \\
        Percentage & 99.67\% & 0.25\% & 0.06\% & 0.02\% \\
        \bottomrule
    \end{tabular}
    }
    \supparxiv{\vspace{-1mm}}{}
    \caption{Model counts per architecture in the training set. The generators are predominantly latent diffusion models.}
    \label{tab_apdx:modelcounts}
\end{table}

\begin{figure}[t]
    \centering
    \includegraphics[width=\supparxiv{0.4\linewidth}{0.4\linewidth}]{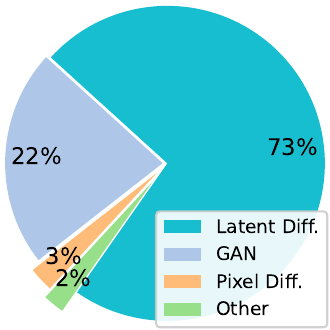} %
    \caption{Number of images per generator type in the training set.}\supparxiv{\figvspace{}}{}
    \label{fig_apdx:imagecounts}
\end{figure}

\secVspace{}
\section{Training settings}
\label{appendix:training_hyperparams}
\secVspace{}

For training our classifiers, we use \texttt{AdamW} optimizer~\cite{loshchilov2018decoupled} with a learning rate of \texttt{2e-5}, a weight decay of \texttt{1e-2}, a batch size of \texttt{512}, and mixed precision~\cite{micikevicius2018mixed}. 
We use a cosine weight decay with a warmup of 20\% of the total iterations. We train our models for \texttt{52K} iterations using this setting. 
For the models in \supparxiv{Figures {\color{red}1} and {\color{red}4}}{\Cref{fig:nummodels,fig:ablations_numImg}}, we employ shorter training iterations (\texttt{3K}) due to the computational overhead associated with training a substantial number of models for statistical analysis.
We chose this number of iterations since we found that classifier performance begins to plateau with approximately this amount of training (\Cref{fig_apdx:ablations_numitr}). %

\begin{figure}[t]
    \centering
    \adjustbox{max width=0.95\linewidth}{
    \begin{tabular}{lrrrrr}
        \toprule
         & Commercial & Latent Diff. & GAN & Pixel Diff. & Other \\
        \midrule
        Models & 11 & 6 & 2 & 1 & 1 \\
        Images & 14918 & 6000 & 2000 & 2000 & 1000 \\
        \bottomrule
    \end{tabular}
    }
    \supparxiv{\vspace{-1mm}}{}
    \caption{Evaluation set composition.}\figvspace{}
    \label{fig_apdx:evalset_composition}
\end{figure}

\supparxiv{\vspace{-1mm}}{}

\begin{figure}[t]
    \centering
    \includegraphics[width=\supparxiv{0.5\linewidth}{0.6\linewidth}]{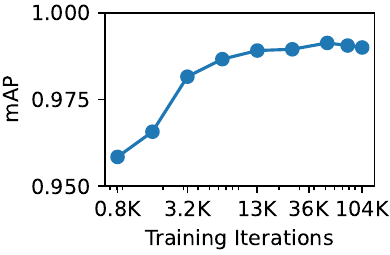}
    \supparxiv{\vspace{-1mm}}{}
    \caption{\textbf{Impact of training iterations.} The performance of the classifier plateaus beyond 3K iterations.}\supparxiv{\figvspace{}}{}
    \label{fig_apdx:ablations_numitr}
\end{figure}

\secVspace{}
\section{Example model project page}
\secVspace{}
\label{appendix:example_model_card}
\supparxiv{\vspace{-3mm}}{}
\begin{figure}[ht]
    \centering
    \includegraphics[width=\linewidth]{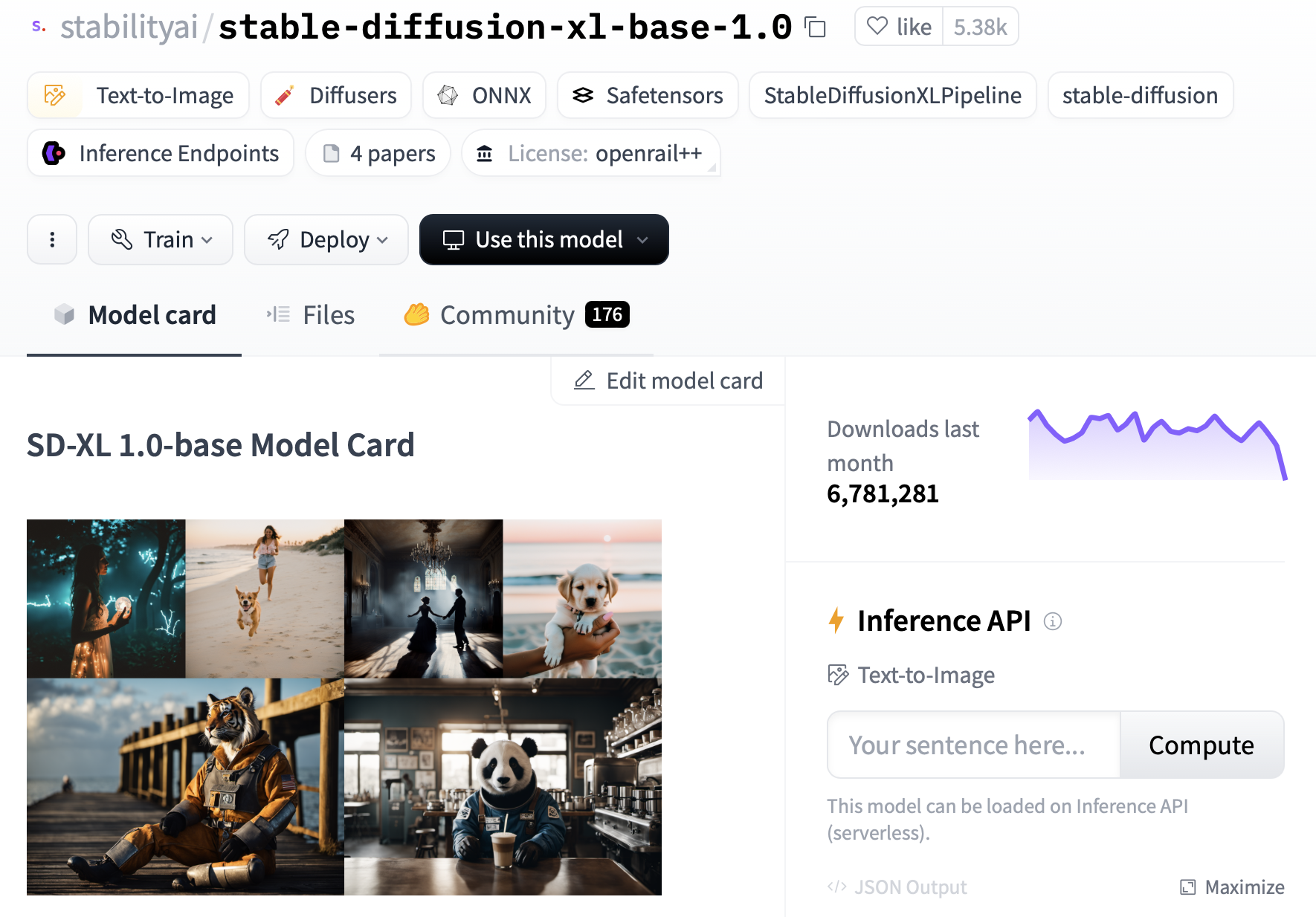}
    \caption{Example model project page from Hugging Face~\cite{HuggingFaceDiffusersWeb}.}\supparxiv{\vspace{-1mm}}{}
    \label{fig_apdx:example_model_card}
\end{figure}

\Cref{fig_apdx:example_model_card} shows a project page from Hugging Face~\cite{HuggingFaceDiffusersWeb}. We can see the tags associated with the model (e.g., Text-to-image, pipeline type, license), number of downloads, and sample images.

}

\end{document}